\documentclass[11pt,a4paper]{article}
\usepackage{times,latexsym}
\usepackage{url}
\usepackage[T1]{fontenc}

\usepackage[acceptedWithA]{tacl2018v2}

\usepackage{xspace,mfirstuc,tabulary}

\newif\iftaclinstructions
\taclinstructionsfalse 
\iftaclinstructions
\renewcommand{\confidential}{}
\renewcommand{\anonsubtext}{(No author info supplied here, for consistency with
TACL-submission anonymization requirements)}
\newcommand{\instr}
\fi

\iftaclpubformat 

\else

\fi

\usepackage{booktabs}
\usepackage{xspace}
\usepackage{tabularx}
\usepackage{xhfill}
\usepackage{xcolor}
\usepackage{multirow}
\usepackage{mdframed}
\usepackage{tikz}
\usepackage{graphicx}
\usepackage{bbm}

\usepackage{pgfplots}
\pgfplotsset{compat=1.15}

\usepackage{mathtools}
\usepackage[colorinlistoftodos,prependcaption,textsize=tiny]{todonotes}
\usepackage{xargs}

\definecolor{forestgreen}{HTML}{009B55}
\definecolor{sepia}{HTML}{671800}
\definecolor{midnightblue}{HTML}{006795}
\definecolor{orangered}{HTML}{E24C00}
\definecolor{bblue}{HTML}{4F81BD}
\definecolor{rred}{HTML}{C0504D}
\definecolor{ggreen}{HTML}{9BBB59}
\definecolor{ppurple}{HTML}{9F4C7C}

\newcommand\gold{\textsc{Gold}\xspace}

\newcommand\pegasus{\textsc{Pegasus}\xspace}
\newcommand\frost{\textsc{Frost}\xspace}
\newcommand\frostecp{\textsc{Frost} (\textsc{ecp})\xspace}
\newcommand\frostecpp{\textsc{Frost} (\textsc{ecpp})\xspace}

\newcommand\entfscore{\textsc{EntF1}\xspace}
\newcommand\entprec{\textsc{EntPrec}\xspace}
\newcommand\rouge{\textsc{rouge}\xspace}

\newcommand\rougetwo{\textsc{rouge-2}\xspace}
\newcommand\rougel{\textsc{rouge-l}\xspace}

\title{Planning with Learned Entity Prompts for Abstractive Summarization}

\author{
Shashi Narayan \\ Google Research \\ \texttt{\small shashinarayan@google.com} \And
Yao Zhao \\ Google Brain \\ \texttt{\small yaozhaoyz@google.com} \And 
Joshua Maynez \\ Google Research \\ \texttt{\small joshuahm@google.com} \AND 
Gon\c{c}alo Simoes \\ Google Research \\ \texttt{\small gsimoes@google.com} \And
Vitaly Nikolaev \\ Google Research \\ \texttt{\small vitalyn@google.com} \And 
Ryan McDonald\thanks{$\;\;$Work done while Ryan was at Google.} \\ ASAPP \\ \texttt{\small ryanmcd@asapp.com}
}

\date{}

\begin{document}
\maketitle
\begin{abstract}

We introduce a simple but flexible mechanism to learn an intermediate plan to ground the generation of abstractive summaries. Specifically, we prepend (or \emph{prompt}) target summaries with entity chains -- ordered sequences of entities mentioned in the summary. Transformer-based sequence-to-sequence models are then trained to generate the entity chain and then continue generating the summary conditioned on the entity chain and the input. We experimented with both pretraining and finetuning with this content planning objective. 
When evaluated on CNN/DailyMail, XSum, SAMSum and BillSum, we demonstrate empirically that the grounded generation with the planning objective improves entity specificity and planning in summaries for all datasets, and achieves state-of-the-art performance on XSum and SAMSum in terms of \rouge. Moreover, we demonstrate empirically that planning with entity chains provides a mechanism to control hallucinations in abstractive summaries. By prompting the decoder with a modified content plan that drops hallucinated entities, we outperform state-of-the-art approaches for faithfulness when evaluated automatically and by humans.
\end{abstract}

\section{Introduction}

\newcite{jones1993} described text summarization -- the task of generating accurate and concise summaries from source document(s) -- as a three-step process: (i) Building the source representation from the source document(s), (ii) Learning a summary representation from the source representation and (iii) Synthesizing the output summary text. Common to most traditional methods, an input representation was learned by semantically analyzing the source text, the summary representation was then learned by modifying and refining the input representation, and finally the summary was generated grounded to the intermediate summary representation \cite{luhn58,muc:summarization,barzilay-elhadad-1997-using,mihalcea-tarau-2004-textrank}.

State-of-the-art neural summarizers are powerful representation learners and conditional language models, thanks to sequence-to-sequence architectures (seq2seq) with attention and copy mechanism \cite{lstm,Bahdanau2015NeuralMT,see-acl17}, Transformer architectures with multi-headed self-attention \cite{transformer}, and large pretrained conditional language models \cite{unilm_arxiv19,mass_icml19,bart,rothe2020leveraging,t5,zhang2019pegasus}. However, the grounding of summary generation that was inherent to most traditional methods is yet to be achieved in neural summarization. The attention mechanism \cite{Bahdanau2015NeuralMT}, especially in pretrained encoder-decoder models \cite{bart,t5,zhang2019pegasus}, plays a key role in aligning summary content to the input, yet undesired hallucinations are common in generated summaries \cite{maynez-etal-2020-faithfulness,kryscinski-etal-2020-evaluating,gabriel2020figure}.

\begin{figure*}[t!]
    \centering
    \begin{tabular}{ c c } 
    \includegraphics[scale=0.305]{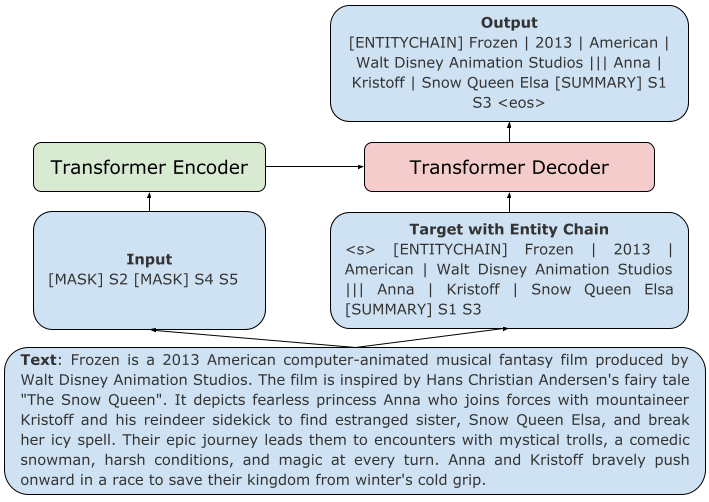} & 
     \includegraphics[scale=0.305]{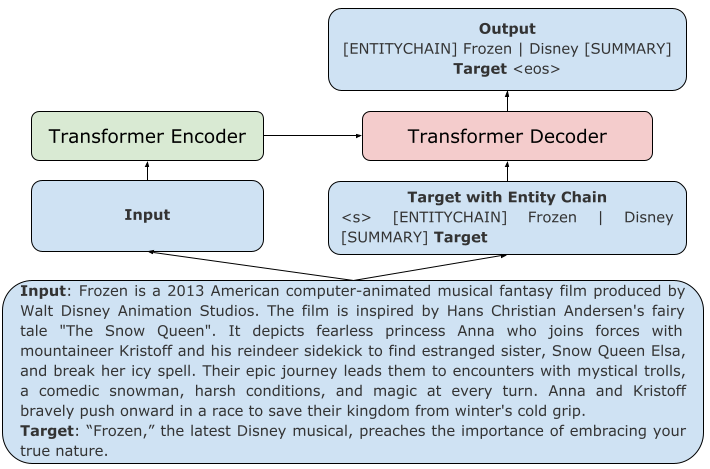}
    \end{tabular}
    \vspace{-0.5cm}
    \caption{Pretraining and finetuning for abstractive summarization with entity chains.}
    \label{fig:frostobjective}
    \vspace{-0.6cm}
\end{figure*}

In this paper, we investigate {\em Entity Chains} -- ordered sequences of entities\footnote{We use the term `entity' broadly and consider named entities, dates and numbers to form an entity chain.} in the summary -- as an intermediate summary representation to better plan and ground the generation of abstractive summaries.  
During training, we construct an augmented target summary by extracting and prepending its corresponding entity chain (Figure~\ref{fig:frostobjective}, right).
At test time, the model must generate both the entity chain followed by the summary.
Concretely, we use Transformer-based encoder-decoder \cite{transformer} models;  a transformer encoder first encodes the input and a transformer decoder generates (i) an intermediate summary representation in the form of an entity chain; and (ii) the summary conditioned on the entity chain and the input. We evaluate our approach on 
four popular summarization datasets: CNN/DailyMail highlight generation \cite{hermann-nips15}, XSum extreme summarization \cite{narayan-etal-2018-xsum}, SAMSum dialogue summarization \cite{gliwa-etal-2019-samsum} and BillSum \cite{kornilova-eidelman-2019-billsum}, and show that the state-of-the-art \pegasus \cite{zhang2019pegasus} pretrained models finetuned with the planning objective clearly outperform regular finetuning in terms of entity specificity and planning in generated summaries on all datasets. 
We further demonstrate that this simple planning mechanism can be  easily used for pretraining summarization models to do entity-level content planning and summary generation. Similar to \pegasus pretraining, we mask important sentences from an input document, extract an entity chain from the masked sentences, and generate these gap-sentences prepended with their entity chain from the rest of the document (Figure~\ref{fig:frostobjective}, left). We see further gains with pretraining achieving state of the art performance on XSum in terms of \rouge. 
We further demonstrate how the entity-level content planning in summarization can be easily leveraged to mitigate hallucinations in abstractive summaries. In particular, we modify the predicted entity chain to only keep entities that are seen in the document and then generate the summary prompted with the modified entity chain, outperforming state-of-the-art approaches when evaluated automatically and by humans for faithfulness. Our main contributions are as follows: 

\paragraph{Planned and Grounded Abstractive Summarization} 
We introduce a novel training objective to neural summarization models for content planning with entity chains, and study integrating it with supervised finetuning and self-supervised pretraining objectives without altering the models themselves. As the entity chains are extracted from the reference summaries during training, our models learn to ground the generation of  summaries to the entity chains found in them. Hence, we refer to this objective by \frost for its ability to try to ``FReeze entity-level infOrmation in abstractive SummarizaTion with planning.'' 

\paragraph{Controlled Abstractive Summarization with Entity Chains}
\frost provides a very effective knob for entity-level content modification in abstractive summaries. In this paper we empirically demonstrate how \frost is critical for faithfulness by enabling the {\em drop-prompt} mechanism where we drop out hallucinated entities from the predicted content plan and prompt the decoder with this modified plan to generate faithful summaries.
We further qualitatively demonstrate that \frost enables generation of summaries (i) with topical diversity by choosing different sets of entities from the source to plan what we want to discuss in the summary, and (ii) with style diversity by reordering entities in the predicted plan to get an equivalent summary but with a different entity emphasis.

\section{Related Work}

\paragraph{Content Planning for Summarization.}
Traditional methods argue on the granularity of linguistic, domain and communicative information included in the source representation needed in order to plan and build better summary representations. Some argued to use a deep semantic analysis of the source text, such as Rhetorical Structure Theory (RST; \citeauthor{rst}, \citeyear{rst}) or MUC-style representations \cite{muc:summarization} to interpret the source texts, while others used a shallow semantic analysis using only word frequency \cite{luhn58} or lexical chains \cite{barzilay-elhadad-1997-using}. 

Recent encoder-decoder models for text generation \cite{Bahdanau2015NeuralMT,NIPS2014_5346,transformer,rothe-etal-2020-leveraging,bart,t5,zhang2019pegasus} tend to perform text generation in an end-to-end setting, which means that most approaches do not explicitly model content planning. \newcite{wiseman-etal-2018-learning} and \newcite{hua-wang-2020-pair} 
start the generation process by building templates which are then used for realization. In data-to-text generation, \newcite{Puduppully_Dong_Lapata_2019} generate a content plan highlighting which information should be mentioned in the table and in which order.
In story generation, there has been some work on exploring events \cite{DBLP:conf/aaai/MartinAWHSHR18} or sequences of words \cite{DBLP:conf/aaai/YaoPWK0Y19} to plan ahead when creating a consistent story. We are not aware of any similar work on content planning for summarization using encoder-decoder models.

\paragraph{Pretraining for Summarization and Planning.}

Pretrained transformer-based models has dramatically changed the text generation space and summarization is no exception to this. Most models focus on task-agnostic pretraining using the left-to-right language modelling objective \cite{gpt,Khandelwal_2019,unilm_arxiv19} or reconstructing the corrupted input text using a sequence-to-sequence framework \cite{mass_icml19,bart,t5}. There has been few attempts towards task-specific pretraining for summarization to teach models to do better content selection. \newcite{zhang2019pegasus} proposed to select important sentences from an input document as a proxy for human-authored summary and then to generate them from the rest of the document. \newcite{qurious} proposed question generation pretraining to better align with summarization. To the best of our knowledge we are the first one to propose a solution that incorporates content planning directly into pretraining.

\paragraph{Controlled Abstractive Summarization.}

There is a growing interest in enabling users to specify high level characteristics such as length, keywords, topic, in order to generate summaries that better suit their needs. In most cases, these features are first manually provided or estimated using third-party content selectors, and then either (i) encoded along with the input \cite{Fan_2018,he2020ctrlsum,dou2020gsum} or (ii) used to filter beams for lexically constrained decoding \cite{mao2020constrained}, to control the summary generation. On the contrary, 
our approach is more generic as we do not rely on external systems or data to augment the input; users can prompt the decoder with a desired content plan in the form of an entity chain to control the summary. 

\section{Content Planning with Entity Chains}
\label{sec:contentplan-with-ec}

We introduce a new training objective for encoder-decoder generative models to do content planning while summarizing.

\paragraph{Model Formulation.} Let $d$ be an input document, we aim to teach our model to first generate a content plan $c$ for the summary $s$ as $p(c|d)$, and then generate the summary $s$ as $p(s|c,d)$. We define the ordered chain of entities observed in the summary $s$ as its content plan. Instead of modeling $p(c|d)$ and $p(s|c,d)$ separately, we take a simpler approach, we train an encoder-decoder model to encode the document $d$ and generate the concatenated content plan and summary sequences $c;s$, essentially the decoder first predicts the entity chain $c$ and then continues predicting the summary $s$ using both $c$ and $d$. We prefix $c$ and $s$ with special markers ``{\footnotesize [ENTITYCHAIN]}'' and ``{\footnotesize [SUMMARY]}'', respectively, as shown in Figure~\ref{fig:frostobjective}. If $s$ consists of multiple sentences, we use sentence markers ``$|||$'' to mark them in $c$. The model is trained with the standard maximum-likelihood objective generating the augmented target $c;s$.

\paragraph{Pretraining Content Plans.}

We modified \pegasus \cite{zhang2019pegasus} to pretrain our models for entity-level content planning and summary generation.\footnote{We experimented with \pegasus, but our technique can be used with any pretraining objectives that require sentence-level input corruptions.}
In particular, we select a maximum of $n$ most important sentences using self-\rouge from  an input document, the selected sentences work as a proxy for a human-authored abstractive summaries for the rest of the document.
We construct a target by prepending the selected sentences dynamically with their entity chain. Our model is then trained to generate this target from the rest of the document.\footnote{The \pegasus objective uses a summary-document length ratio to select $n$. This could lead to an undesirably long summary when the input document is very long. Modeling such summaries prepended with long entity chains effectively is beyond the limit of our decoders (256 sentencepieces). Hence, we set $n=5$.}

\paragraph{Modeling Entity-Level Lexical Cohesion and Coherence.}
As entities in the summary contribute to the continuity of lexical meaning of the summary, we hypothesize that by learning to predict the entity chain $c$ in advance, we enforce our model to learn entity-level lexical cohesion \cite{barzilay-elhadad-1997-using} and coherence \cite{Halliday76a,azzam-etal-1999-using} in the summary. We hope that by doing so our model will be better at predicting pertinent entities ({\em entity specificity}) in their right order ({\em entity planning}) in generated summaries; the prediction of entities in the entity chain $c$ (as $p(c|d)$ in \frost) will be less susceptible to local correlations compared to when predicting them directly in the summary $s$ (as $p(s|d)$). Furthermore, as $c$ is predicted in advance and the generation of $s$ is grounded to $c$,\footnote{Here, $s$ is not strictly constrained to the entity chain $c$. We hope that this will happen given $c$ is extracted from $s$ during the training time. Future work will focus on constraining $s$ to $c$, e.g., using a checklist model \cite{kiddon-etal-2016-globally} or entity-chain constrained decoding \cite{mao2020constrained}.} our model will be better equipped to predict correct events relating to different entities in $c$ with full access to $c$ and not just entities to the left, already decoded.

\begin{table*}[t!]
  \begin{center}{\scriptsize 
  \begin{tabular}{ l | c | c | c c c c | c c c } 
    \toprule
    \multirow{2}{*}{Dataset} & \multirow{2}{*}{Size} & \multirow{3}{*}{Case} & \multicolumn{7}{c}{Target Summaries (Validation)} \\
    & & & avg. & avg. & avg. & \% target & \multicolumn{3}{c}{total} \\ 
    & train/dev/test & & sent. & ent. & uniq. ent. & (no ent.) & named & date & number \\ \midrule 
    BillSum & 18.9k/--/3.3k & cased & 4.38 & 14.78 & 11.10 & 0.18 & 28931 & 2578 & 16646 \\
    CNN/DailyMail &  287k/13.4k/11.5k & cased & 4.11 & 7.55 & 6.92 & 0.10 & 74292 & 3569 & 23094 \\
    SAMSum & 14.7k/818/819 & cased & 2.03 & 3.59 & 3.01 & 0.37 & 2594 & 33 & 309 \\
    XSum &  204k/11.3k/11.3k & cased & 1.00 & 2.81 & 2.80 & 5.97 & 24682 & 777 & 6287 \\
    \bottomrule
  \end{tabular}}
  \end{center}
  \vspace{-0.3cm}
  \caption{Abstractive summarization datasets studied in this work. We report on their train/validation/test sizes and how they were processed (cased/uncased). To better understand the effect of summary planning with entity chains, we report on average number of sentences (avg. sent.), average number of entities (avg. ent.) and average number of unique entities (avg. uniq. ent.), per target in validation sets. We also report on total number of named entities, date and number in target summaries.} \label{table:datasets-stat}
  \vspace{-0.4cm}
\end{table*}

\paragraph{Controlled Generation with Entity Prompts.}

An advantage of training to generate the summary $s$ following the generation of the plan $c$ using the same decoder is that now during the inference time the decoder can be easily prompted with any desired content plan $c'$ to control the content in the output summary $s'$ with a probability of $p(s'|c',d)$. In Section~\ref{sec:results}, we prompt our decoder with modified content plans to mitigate hallucinations and to generate diverse summaries.

\section{Experimental Setup}
\label{sec:setup}

\subsection{Base and Large Models}

We experiment with both base and large transformer architectures \cite{transformer}. The base architecture has $L=12$, $H=768$, $F=3072$, $A=12$ (223M parameters) and the large architecture has $L=16$, $H=1024$, $F=4096$, $A=16$ (568M parameters), where $L$ denotes the number of layers for encoder and decoder Transformer blocks, $H$ for the hidden size, $F$ for the feed-forward layer size, and $A$ for the number of self-attention heads. All pretrainings are done with a batch size of 1024, whereas all finetuning experiments are done with a smaller batch size of 256. 
For optimization, we use Adafactor \cite{adafactor} with square root learning rate decay and dropout rate of $0.01$ during pretraining and $0.0001$ during finetuning. All finetuned models were decoded with a beam size of $8$ and a length-penalty of $0.8$.\footnote{
We will release all models and other resources for easy replication of our results.} 

\subsection{Datasets and Entity Annotations}

\paragraph{Pretraining Datasets.} Following \newcite{zhang2019pegasus}, our model pretraining also relied on two large web corpus which were processed to look like plain text: (i) \textbf{C4} \cite{t5} is composed of 350M Web-pages that were obtained from Common Crawl, and (ii) \textbf{HugeNews} \cite{zhang2019pegasus} is composed of 1.5B news and news-like articles from 2013-2019. This dataset includes articles from multiple allowlisted sources including news publishers, high-school newspapers and blogs. 

\paragraph{Abstractive Summarization Datasets.}

We evaluate our models on four summarization datasets: CNN/DailyMail highlight generation \cite{hermann-nips15}, XSum extreme summarization \cite{narayan-etal-2018-xsum}, SAMSum dialogue summarization \cite{gliwa-etal-2019-samsum} and BillSum summarizing US Congressional bills \cite{kornilova-eidelman-2019-billsum}. We use the publicly available versions through the TFDS Summarization Datasets\footnote{https://www.tensorflow.org/datasets/catalog}. We use the original train/validation/test splits for them. For BillSum where the validation split was not provided, we split 10\% of the training set to serve as validation. Inputs and outputs were truncated to 512 and 128 for XSum and SAMSum, and, 1024 and 256 for CNN/DailyMail and BillSum.  Table~\ref{table:datasets-stat} provides more insights into these datasets to better understand the effect of summary planning with entity chains.

\paragraph{Entity Chain Annotation.} 

We experimented with entity chains consisting of named entities, dates and numbers.
We annotate the whole document in the pretraining datasets to allow dynamic construction of summaries and their entity chains during pretraining. We only annotate the reference summaries for the finetuning summarization datasets. We use a BERT-based tagger trained on CoNLL-2003 data
\cite{tjong-kim-sang-de-meulder-2003-introduction} to identify named entities, and regular expressions to identify dates and numbers \cite{guu2020realm}. See Table~\ref{table:datasets-stat} for the number of named entities, dates and numbers found in different datasets. 

Table~\ref{table:pretrainparameters} and~\ref{table:finetuneparameters} in Appendix~\ref{app-sec:hyperparameters} present hyperparameters used for pretraining and finetuning \pegasus and \frost base and large sized models. 
We used Cloud TPU v3 accelerators for training.

\subsection{Evaluation Measures}

We evaluate our \frost models on \rouge, {\em entity specificity}, {\em entity planning}, {\em faithfulness} using automatic and human evaluations, and {\em overall quality} by humans. Our models predict a summary plan in the form of an entity chain, followed by a summary. All evaluations are done on the summary, the predicted entity chains are stripped out.

\paragraph{Summary-level \rouge.} We report \rouge F1 scores \cite{rouge} to assess generated summaries.\footnote{We lowercased candidate and reference summaries and used \texttt{pyrouge} with parameters ``-a -c 95 -m -n 4 -w 1.2.''} 

\paragraph{Entity Planning.} We evaluate the quality of content plans learned by our models by assessing the entities and the order in which they appear in the summary. We annotate both predicted and reference summaries with named entities, dates and numbers, and report on \rouge F1 scores for entity chains found in the predicted summaries against corresponding reference entity chains.

\paragraph{Entity Specificity.}  We compare entities in predicted and reference summaries, and report average entity F1-scores (\entfscore) in predicted summaries. We lower-case and remove duplicate entities; we report on the exact match with reference entities.

\paragraph{Faithfulness.}

For entity-level faithfulness, we report on \entprec, measuring the precision of entities in predicted summaries against their input documents. \entfscore, in comparison, evaluates predicted summaries for entity specificity against reference summaries and is not a measure for faithfulness.
For summary-level faithfulness, we follow \newcite{maynez-etal-2020-faithfulness} and report on textual entailment \cite{Pasunuru-multireward18,falke-etal-2019-ranking,Kryscinski2019EvaluatingTF}. In particular, we report the probability of a summary entailing ({\em Entail.}) its input document using an entailment classifier trained by fine-tuning an uncased BERT-Large pretrained model \cite{bert} on the Multi-NLI dataset \cite{mnli}.

We further assess summary faithfulness by humans. Our annotators, proficient in English, were tasked to read the document carefully and then grade its summary on a scale of 1-5 ({\em fully unfaithful}, {\em somewhat unfaithful}, {\em 50-50}, {\em somewhat faithful} and {\em fully faithful}); a summary is ``fully faithful'' if all of its content is fully supported or can be inferred from the document. 

\paragraph{Overall Summary Quality.} Finally we assess the overall quality of summaries by humans. We ask our annotators to read the document carefully and then grade its summary on a scale of 1-5 ({\em poor summary}, {\em better than poor}, {\em okay summary}, {\em better than okay} and {\em great summary}). To improve the annotator agreement, we clearly define 4 features that a ``great summary'' should have: {\em Relevant} (summary should have most important information from the document), {\em Accurate} (summary should be accurate with respect to the document or the expert knowledge),\footnote{With {\em accurate} we mean {\em factual} to the background knowledge and not just {\em faithful} to the document; as it is natural to construct summaries that integrate with the author's background knowledge \cite{maynez-etal-2020-faithfulness}.} {\em Concise} (summary should not have redundant or less-important content), and {\em Fluent} (summary should be grammatically correct and coherent). A {\em great summary} should pass on all 4 features, a {\em better than okay}  should have 3 out of 4 features, and so on.

For both human assessments, we collected 3 ratings for each (document, summary) pair and report the average of all assigned labels (1-5) to a system. We conducted 3 pilot studies for each setup to train our annotators with examples to improve their understanding of the task. Additionally, extra measures were taken to improve agreements among our annotators. For example, for the faithfulness assessment, when one of {\em somewhat unfaithful}, {\em 50-50} and {\em somewhat faithful} were selected, annotators were asked to also specify what was faithful or unfaithful in the summary. Similarly for the overall quality assessment, when one of {\em better than poor}, {\em okay summary} and {\em better than okay} were selected, they were asked to list all features on which the candidate summary fails. Figure~\ref{fig:faithfulness} and~\ref{fig:overall} in Appendix~\ref{app-sec:human-eval} show detailed instructions for human evaluations for faithfulness and overall quality of summaries, respectively.



\section{Results}
\label{sec:results}

\subsection{\frost Ablations}
\label{subsec:ablations}

\begin{figure}[t!]
  \center{\scriptsize  
  \setlength\tabcolsep{0.1cm}
    \begin{tabular}{p{7.5cm}}
    \toprule 
    \textbf{Sentence-level Entity Chain with Target}: \textcolor{midnightblue}{[EntityChain]} \textcolor{sepia}{Frozen} $|$ \textcolor{sepia}{Disney} \textcolor{midnightblue}{[Summary]} ``\textcolor{sepia}{Frozen},'' the latest \textcolor{sepia}{Disney} musical, preaches the importance of embracing your true nature. \textcolor{midnightblue}{[EntityChain]} \textcolor{orangered}{Princess Anna} $|$ \textcolor{orangered}{Kristoff} $|$ \textcolor{orangered}{Snow Queen Elsa} \textcolor{midnightblue}{[Summary]} It depicts fearless \textcolor{orangered}{Princess Anna} who joins forces with mountaineer \textcolor{orangered}{Kristoff} and his reindeer sidekick to find estranged sister, \textcolor{orangered}{Snow Queen Elsa}, and break her icy spell. \\
    \midrule
    \textbf{Summary-level Entity Chain with Target}: \textcolor{midnightblue}{[EntityChain]} \textcolor{sepia}{Frozen} $|$ \textcolor{sepia}{Disney} $|||$ \textcolor{orangered}{Princess Anna} $|$ \textcolor{orangered}{Kristoff} $|$ \textcolor{orangered}{Snow Queen Elsa} \textcolor{midnightblue}{[Summary]} ``\textcolor{sepia}{Frozen},'' the latest \textcolor{sepia}{Disney} musical, preaches the importance of embracing your true nature. It depicts fearless \textcolor{orangered}{Princess Anna} who joins forces with mountaineer \textcolor{orangered}{Kristoff} and his reindeer sidekick to find estranged sister, \textcolor{orangered}{Snow Queen Elsa}, and break her icy spell. \\
    \midrule
    \textbf{Original Target}: ``\textcolor{sepia}{Frozen},'' the latest \textcolor{sepia}{Disney} musical, preaches the importance of embracing your true nature. It depicts fearless \textcolor{orangered}{Princess Anna} who joins forces with mountaineer \textcolor{orangered}{Kristoff} and his reindeer sidekick to find estranged sister, \textcolor{orangered}{Snow Queen Elsa}, and break her icy spell. \\
    \bottomrule
    \end{tabular}     
  }
  \vspace{-0.3cm}
  \caption{An example of sentence-level and summary-level entity chains along with its reference summary.}
  \label{fig:ex-sent-sum-entitychain}
  \vspace{-0.3cm}
\end{figure}

\begin{figure}[t!]
  \footnotesize
  \center{
\begin{tikzpicture}
    \begin{axis}[
        width  = 0.96*\linewidth,
        height = 4cm,
        major x tick style = transparent,
        ybar=2*\pgflinewidth,
        bar width=7pt,
        ymajorgrids = true,
        ylabel = {Fscores},
        symbolic x coords={RL-Sum,R2-EPlan,\entfscore},
        xtick = data,
        ytick = {40,45,50,55},
        scaled y ticks = false,
        enlarge x limits=0.25,
        ymin=35,
        legend style={at={(0.5,1.01)}, anchor=south,legend columns=-1,font=\scriptsize}
    ]
        \addplot[style={bblue,fill=bblue,mark=none}]
            coordinates {(RL-Sum, 41.99) (R2-EPlan, 47.13) (\entfscore,50.66)};

        \addplot[style={rred,fill=rred,mark=none}]
             coordinates {(RL-Sum,41.74) (R2-EPlan, 48.77) (\entfscore,52.11)};

        \addplot[style={ggreen,fill=ggreen,mark=none}]
             coordinates {(RL-Sum,42.69) (R2-EPlan, 50.48) (\entfscore,53.82)};

        \legend{None,Sent-Level,Sum-Level}
    \end{axis}
\end{tikzpicture}
  }
  \vspace{-0.4cm}
  \caption{Sentence-level vs summary-level entity chains. We report summary-level \rougel (RL-Sum), entity chain-level \rougetwo (R2-EPlan) and \entfscore on the CNN/DailyMail validation set. Similar observations were made for other measures. 
  \label{fig:ablation-sum-sent-level}}
  \vspace{-0.4cm}
\end{figure}
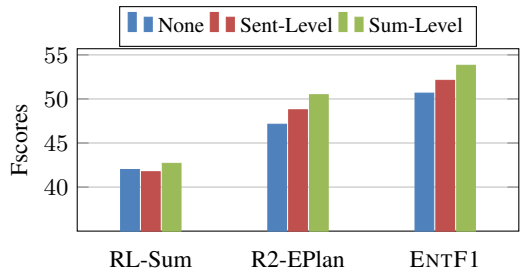

\paragraph{Sentence-Level vs Summary-Level Planning.}

Let the summary $s$ consists of $m$ sentences $s_1 \ldots s_m$ and $c_i$ be the entity chain for the sentence $s_i$, we consider generating the summary $s$ in two ways. {\em Sentence-level} approach trains a model to generate $s$ by consecutively generating the sentence-level content plan $c_i$ followed by its summary sentence $s_i$ with a probability  $p(c_i s_i | c_1 s_1 \ldots c_{i-1} s_{i-1}; d)$; $d$ is the input document. {\em Summary-level} approach first generates a summary-level content plan $c = c_1 ||| \ldots ||| c_m$ and then continue generating the summary $s$ with a probability $p(s|c;d)$; $|||$ are sentence markers.  
The summary-level planning is arguably better suited for summarization than the sentence-level planning. By planning the whole summary beforehand, the summary-level planner would be (i) less susceptible to local correlations than the sentence-level planning while generating the entities, and (ii) a better microplanner in deciding sentence boundaries and avoiding verbose summaries \cite{reiter_dale_1997}. See examples for sentence-level and summary-level entity chains in Figure~\ref{fig:ex-sent-sum-entitychain}.

We finetuned our large models initialized with the \pegasus checkpoint on the CNN/DailyMail dataset for sentence-level vs summary-level entity chain ablations. We did not do this study on the XSum dataset as the XSum summaries are single sentences, which means that sentence-level and summary-level entity chains are the same. In contrast, the CNN/DailyMail summaries consists of multi-sentence highlights. Results are presented in Figure~\ref{fig:ablation-sum-sent-level}.

We found that the summary-level planning is not only superior to the sentence-level planning, it helps with generating better quality summaries in terms of summary-level \rouge (RL-Sum), entity planning (R2-EPlan) and entity specificity (\entfscore). For the rest of the pretraining or finetuning experiments, we focused on summary-level planning unless specified otherwise. 

\pgfplotstableread[row sep=\\,col sep=&]{
  ModelID & R2Sum & R2EPlan & EntAcc \\ 
  5 & 22.18 & 37.39 & 53.11 \\
  10 & 22.08 & 37.66 & 53.25 \\
  15 & 22.32 & 37.76 & 53.30 \\
}\basescores

\begin{figure}[t!]
  \footnotesize
  \center{
    \begin{tikzpicture}
      \begin{axis}[
        name=plot1,
        width=3.0in,
        height=1.1in,
        title={},
        xmin=3, xmax=17,
        ymin=22, ymax=22.4,
        xtick={5, 10, 15},
        ytick={22.0, 22.1, 22.2, 22.3, 22.4},
        xticklabels={\pegasus, \textsc{Frost}(\textsc{F}), \textsc{Frost}(\textsc{P+F})},
        legend pos=outer north east,
        xmajorgrids=false,
        ymajorgrids=true,
        grid style=dashed,
        legend style={at={(0.5,1.01)}, anchor=south,legend columns=-1,font=\scriptsize},
        tick label style={font=\scriptsize},
        ]
        \addplot [bblue] table[x=ModelID,y=R2Sum]{\basescores};
        \legend{R2-Summary}
      \end{axis}
      \begin{axis}[
        name=plot2,
        at=(plot1.below south east), anchor=above north east,
        width=3.0in,
        height=1.1in,
        title={},
        xmin=3, xmax=17,
        ymin=37.3, ymax=37.8,
        xtick={5, 10, 15},
        ytick={37.3, 37.4, 37.5, 37.6, 37.7, 37.8},
        xticklabels={\pegasus, \textsc{Frost}(\textsc{F}), \textsc{Frost}(\textsc{P+F})},
        legend pos=outer north east,
        xmajorgrids=false,
        ymajorgrids=true,
        grid style=dashed,
        legend style={at={(0.5,1.01)}, anchor=south,legend columns=-1,font=\scriptsize},
        tick label style={font=\scriptsize},
        ]
        \addplot [rred] table[x=ModelID,y=R2EPlan]{\basescores};
        \legend{R2-EntityPlan}
      \end{axis}
      \begin{axis}[
        name=plot3,
        at=(plot2.below south east), anchor=above north east,
        width=3.0in,
        height=1.1in,
        title={},
        xmin=3, xmax=17,
        ymin=53, ymax=53.4,
        xtick={5, 10, 15},
        ytick={53.0, 53.1, 53.2, 53.3, 53.4},
        xticklabels={\pegasus, \textsc{Frost}(\textsc{F}), \textsc{Frost}(\textsc{P+F})},
        legend pos=outer north east,
        xmajorgrids=false,
        ymajorgrids=true,
        grid style=dashed,
        legend style={at={(0.5,1.01)}, anchor=south,legend columns=-1,font=\scriptsize},
        tick label style={font=\scriptsize},
        ]
        \addplot [ggreen] table[x=ModelID,y=EntAcc]{\basescores};
        \legend{\entfscore}
      \end{axis}
    \end{tikzpicture}
  }
  \vspace{-0.3cm}
  \caption{
  Finetuning results on the XSum validation set using one of the base-sized pretrained models: \pegasus, \textsc{Frost}(\textsc{F}) and \textsc{Frost}(\textsc{P+F}). All pretrained models were trained for 1.5m steps. See text for more details. We only report on a subset of measures, similar observations were made for other measures. 
  \label{fig:base-model-performance}}
  \vspace{-0.4cm}
\end{figure}
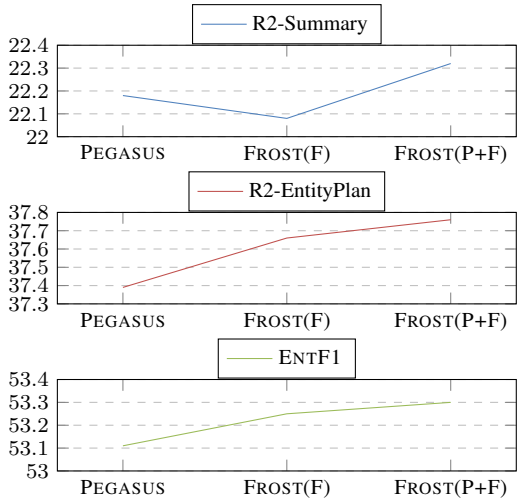

\paragraph{Pretraining for Planning from Scratch.}

We pretrained three base-sized models from scratch: \pegasus-base pretrained with the original gap-sentence objective \cite{zhang2019pegasus} for 1.5m steps, \textsc{Frost}(\textsc{F})-base pretrained with the gap-sentences prepended with their entity chain for 1.5m steps, and \textsc{Frost}(\textsc{P+F})-base first pretrained with the \pegasus objective for 1m steps and then with the \frost objective for another 500k steps. Maximum input and output lengths were set to 512 and 256 sentencepieces during pretraining, respectively. We finetuned these three models on the XSum dataset. Results are shown in 
Figure~\ref{fig:base-model-performance}. 

First of all, our results confirm that the pretraining for planning is beneficial for summarization; both \textsc{Frost}(\textsc{F}) and \textsc{Frost}(\textsc{P+F}) learned better content plans (in terms of entity chain-level \rouge and \entfscore) than \pegasus without sacrificing the summary quality (in terms of summary-level \rouge) significantly. Interestingly, the \textsc{Frost}(\textsc{P+F}) finetuned model outperformed \textsc{Frost}(\textsc{F}) on all measures confirming that the pretraining for planning and summarization is more effective when started with summarization pretrained models such as \pegasus than when trained jointly from scratch. Hence, for our future pretraining experiments we initialize our model with existing \pegasus checkpoint and continue pretraining for content planning and refer to them as \frost for simplicity.

\pgfplotstableread[row sep=\\,col sep=&]{
  Steps & XSum & CNNDailyMail \\ 
  3 & 24.69 & 22.61 \\
  5 & 25.12 & 22.72 \\
  5.9 & 25.32 & 22.89 \\
  6.3 & 25.01 & 22.80 \\
}\largestepsrtwosum
\pgfplotstableread[row sep=\\,col sep=&]{
  Steps & XSum & CNNDailyMail \\ 
  3 & 41.02 & 50.48 \\
  5 & 41.58 & 50.67 \\
  6 & 41.64 & 50.85 \\
  6.3 & 41.55 & 50.78 \\
}\largestepsrtwoechain
\pgfplotstableread[row sep=\\,col sep=&]{
  Steps & XSum & CNNDailyMail \\ 
  3 & 55.18 & 53.82 \\
  5 & 56.00 & 53.90 \\
  5.9 & 56.18 & 54.07 \\
  6.3 & 55.88 & 54.05 \\
}\largestepsentacc
\begin{figure}[t!]
  \footnotesize
  \center{
    \begin{tikzpicture}
      \begin{axis}[
        name=plot1,
        width=3.0in,
        height=1.3in,
        title={},
        ylabel={R2-Sum},
        xmin=2.8, xmax=6.5,
        ymin=22.3, ymax=25.6,
        xtick={3, 5, 5.9, 6.3},
        ytick={},
        xticklabels={\pegasus, 0.1m, 1m, 1.5m},
        legend pos=outer north east,
        xmajorgrids=true,
        ymajorgrids=true,
        grid style=dashed,
        legend style={at={(0.5,1.01)}, anchor=south,legend columns=-1,font=\scriptsize},
        ]
        \addplot [bblue] table[x=Steps,y=XSum]{\largestepsrtwosum};
        \addplot [rred] table[x=Steps,y=CNNDailyMail]{\largestepsrtwosum};
      \end{axis}
      \begin{axis}[
        name=plot2,
        at=(plot1.below south east), anchor=above north east,
        width=3.0in,
        height=1.3in,
        title={},
        ylabel={\entfscore},
        xmin=2.8, xmax=6.5,
        ymin=53.5, ymax=56.5,
        xtick={3, 5, 5.9, 6.3},
        ytick={},
        xticklabels={\pegasus, 0.1m, 1m, 1.5m},
        legend pos=outer north east,
        xmajorgrids=true,
        ymajorgrids=true,
        grid style=dashed,
        legend style={at={(0.5,1.01)}, anchor=south,legend columns=-1,font=\scriptsize},
        ]
        \addplot [bblue] table[x=Steps,y=XSum]{\largestepsentacc};
        \addplot [rred] table[x=Steps,y=CNNDailyMail]{\largestepsentacc};
      \end{axis}
    \end{tikzpicture}
  }
  \vspace{-0.2cm}
  \caption{Finetuning results on the XSum (in \textcolor{bblue}{blue}) and CNN/DailyMail (in \textcolor{rred}{red}) validation sets at various steps during pretraining \frost-Large. Instead of pretraining from scratch, we start with a \pegasus-Large checkpoint, and continue pretraining for additional 1.5m steps with the planning objective. We report finetuning results for the \pegasus finetuned baseline and our models at 0.1m, 1m and 1.5m steps. 
  \label{fig:large-model-step-performance}}
  \vspace{-0.5cm}
\end{figure}
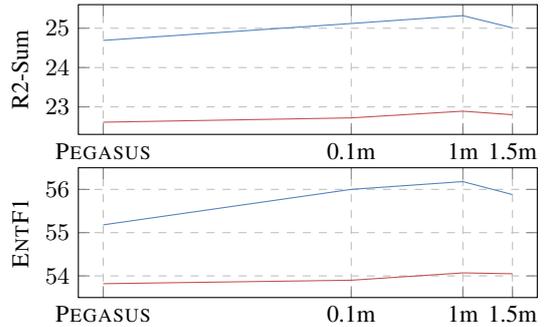

\paragraph{Effect of Pretraining Longer for Planning.}

Based on our findings, we pretrain a large \frost model for planning for summarization starting with an existing \pegasus-Large checkpoint. Figure~\ref{fig:large-model-step-performance} presents results for finetuning these models on the XSum and CNN/DailyMail datasets at various steps during pretraining.

Similar to our findings in Figure~\ref{fig:base-model-performance}, our results with large models further confirm the advantages of the pretraining for planning; improvement over the \pegasus baseline were larger for the XSum dataset than for the CNN/DailyMail dataset. We achieved the best performance on both datasets at the 1m pretraining step. We use the \frost model at 1m pretraining step for our finetuning experiments.

\begin{table}[t!]
  \begin{center}{\tiny 
  \begin{tabular}{ l | c c c } 
    \toprule
    \multirow{2}{*}{\textbf{Model}} & \textbf{Summary} & \textbf{Entity Planning} &  \textbf{Specificity} \\
    & R1/R2/RL & R1/R2/RL & \entfscore \\
    \midrule 
    \multicolumn{4}{c}{\textbf{CNN/DailyMail}} \\
    \midrule 
    \textsc{Lead-3} & 40.31/17.83/36.45 & 57.41/42.32/47.16 & 46.22 \\
    \textsc{Ext-Oracle}$^{\ast}$ & 57.03/34.38/53.12 & 68.79/55.77/59.02 & 59.91 \\
    RoBERTaShare & 39.25/18.09/36.45 & -- & -- \\
    MASS & 42.12/19.50/39.01 & -- & -- \\
    UniLM & 43.33/20.21/40.51 & -- & -- \\
    T5 & 43.52/21.55/40.69 & -- & -- \\
    BART & 44.16/21.28/40.90 & -- & -- \\
    ProphetNet & 44.20/21.17/41.30  & -- & -- \\
    \pegasus & 44.16/21.56/41.30 & -- & -- \\ 
    GSum & \textbf{45.94}/22.32/42.48 & -- & -- \\
    CTRLsum & 45.65/\textbf{22.35}/\textbf{42.50} & 62.19/48.35/52.53 & 51.69 \\
    \hline 
    \pegasus (ours) & 44.05/21.69/40.98 & 61.12/46.46/52.40 & 49.93 \\
    \frostecp & 44.85/21.83/41.80 & \textbf{64.34}/49.85/\textbf{54.98} & 53.17 \\
    \frostecpp & 45.11/22.11/42.01 & 64.28/\textbf{49.86}/54.96 & \textbf{53.22} \\
    \midrule 
    \multicolumn{4}{c}{\textbf{XSum}} \\
    \midrule
    \textsc{Lead-1} & 16.66/1.85/12.26 & 21.45/4.36/19.48 & 5.36 \\
    \textsc{Ext-Oracle}$^{\ast}$ & 28.81/8.61/21.97 & 32.64/13.34/28.85 & 18.78 \\
    RoBERTaShare & 38.52/16.12/31.13 & -- & -- \\
    MASS & 39.75/17.24/31.95 & -- & -- \\
    BART & 45.14/22.27/37.25 & -- & --\\
    GSum & 45.40/21.89/36.67 & -- & -- \\
    \pegasus & 47.60/24.83/39.64 & -- & -- \\ 
    \hline 
    \pegasus (ours) & 47.56/24.87/39.40 & 62.15/39.76/56.36 &  53.48 \\
    \frostecp & 47.44/24.54/39.24 & 63.57/40.45/57.65 & 54.62 \\
    \frostecpp & \textbf{47.80}/\textbf{25.06}/\textbf{39.76} & \textbf{64.09}/\textbf{41.07}/\textbf{58.18} & \textbf{55.49} \\
    \midrule 
    \multicolumn{4}{c}{\textbf{SAMSum}} \\
    \midrule 
    \cite{gliwa-etal-2019-samsum} & 40.99/17.72/38.30 & -- & -- \\ 
    \hline 
    \pegasus (ours) & 52.27/\textbf{28.34}/\textbf{47.83} & 72.42/51.07/64.85 & 81.17 \\
    \frostecp & \textbf{52.39}/27.70/47.82 & 74.42/\textbf{55.32}/66.35 & \textbf{83.60} \\
    \frostecpp & 51.86/27.67/47.52 & \textbf{75.02}/55.19/\textbf{66.80} & \textbf{83.60} \\
    \midrule 
    \multicolumn{4}{c}{\textbf{BillSum}} \\
    \midrule 
    \pegasus & 59.67/41.58/47.59 & -- & -- \\ 
    \hline 
    \pegasus (ours) & 59.33/\textbf{41.60}/54.80 & 69.99/62.79/66.17 & 61.91 \\
    \frostecp & 58.76/40.34/54.03 & 70.84/63.59/66.86 & 62.38 \\
    \frostecpp & \textbf{59.50}/41.17/\textbf{54.85} & \textbf{71.67}/\textbf{64.56}/\textbf{67.79} & \textbf{63.38} \\
    \bottomrule
  \end{tabular}}
  \end{center}
  \vspace{-0.3cm}
  \caption{Final results on abstractive summarization datasets compared with the previous SOTA. We report results from RoBERTaShare \cite{rothe2020leveraging}, MASS \cite{mass_icml19}, UniLM \cite{unilm_arxiv19}, T5 \cite{t5}, BART \cite{bart}, ProphetNet \cite{qi-etal-2020-prophetnet},  \cite{gliwa-etal-2019-samsum}, CTRLsum \cite{he2020ctrlsum}, GSum \cite{dou2020gsum} and \pegasus \cite{zhang2019pegasus}. We also include commonly reported \textsc{Lead-}$n$ baselines (selecting top $n$ sentences from the document) and extractive oracle (\textsc{Ext-Oracle}; selecting best set of sentences from the document with the most overlapping content with its reference summary), for the CNN/DailyMail and XSum datasets. \textsc{Ext-Oracle}s are masked with $\ast$ and are not directly comparable. 
  All results are reported on the publicly available test sets.} \label{table:summarization-results}
  \vspace{-0.6cm}
\end{table}

\subsection{Abstractive Summarization Results}

Table~\ref{table:summarization-results} presents our final results from finetuning our large models on summarization datasets. For each dataset, we first report results from earlier work directly taken from corresponding papers; our results are in the bottom blocks for each dataset. We finetune our own \pegasus using the standard approach (document to summary). We report results from \frost (\textsc{ecp}; Entity Chain Planning), i.e., \pegasus finetuned with the \frost objective (document to content plan and summary). Finally, we report on \frost (\textsc{ecpp}; Entity Chain Planning with Pretraining), our models both pretrained and finetuned with the \frost objective.

We find that even simply finetuning an existing \pegasus pretrained model to do content planning and summarization (as in \frostecp) leads to improvements in entity specificity (\entfscore) and the quality of entity plans in summaries (Entity Planning \rouge), across all datasets. In fact in some cases, better content plans lead to large improvements in summary-level \rouge as well, for example, \frostecp improve on \pegasus from 44.05/21.69/40.98 to 44.85/21.83/41.80 on \rouge scores for CNN/DailyMail summaries. 

The pretraining for content planning and generation in \frostecpp further improves the entity chain quality for CNN/DailyMail, XSum and BillSum, and the summary-level \rouge for CNN/DailyMail and XSum. Our \frost models establish new state-of-the-art \rouge results on XSum. For CNN/DailyMail, we perform inferior to CTRLsum \cite{he2020ctrlsum} and GSum \cite{dou2020gsum} on \rouge scores. However, we outperform CTRLsum on entity planning and specificity. Further discussion on comparing our methods to controlled summarization systems such as CTRLsum can be found in Section~\ref{subsec:enc-guided-control}.


\begin{table*}[t!]
  \begin{center}{\tiny 
  \begin{tabular}{ l | c c c | c c c } 
    \toprule
    \multirow{3}{*}{\textbf{Model}} & \multicolumn{3}{c|}{\textbf{Test set with Extractive Entity Chains Only (3.5k)}} & \multicolumn{3}{c}{\textbf{Test set with Non-Extractive Entity Chains Only (7.8k)}} \\
    & \textbf{Summary} & \textbf{Entity Planning} & \textbf{Specificity} & \textbf{Summary} & \textbf{Entity Planning} & \textbf{Specificity} \\
    & R1/R2/RL & R1/R2/RL & \entfscore & R1/R2/RL & R1/R2/RL & \entfscore \\
    
    \midrule 
    \multicolumn{7}{c}{\textbf{Models trained on the original XSum dataset (204k/11.3k/11.3k)}} \\
    \midrule 
    \pegasus & 44.67/21.75/36.37 & 61.54/28.84/58.02 & 60.88 & 48.85/26.26/40.75 & 62.42/44.64/55.62 & 50.17 \\
    \frostecp & 44.53/21.38/36.28 & 63.60/29.20/60.01 & 62.04 & 48.74/25.95/40.56 & 63.56/45.48/56.60 & 51.31 \\
    \frostecpp & \textbf{45.08/22.01/36.80} & 63.73/30.08/60.10 & 63.03 & \textbf{49.02/26.42/41.08} & \textbf{64.25/45.98/57.32} & \textbf{52.12} \\
    $\;\;\;$($d \rightarrow c_{\text{drop}};s$) & 44.97/21.97/36.77 & \textbf{67.10/30.19/63.58} & 64.41 & 44.93/21.41/37.39 & 48.89/30.58/43.76 & 33.68 \\
    $\;\;\;$($d \rightarrow c_{\text{oracle}};s$)$^{\ast}$ & 50.26/27.33/43.12 & 98.55/53.94/98.53 & 98.36 & 61.80/40.55/56.20 & 99.10/92.10/99.08 & 98.69 \\
    \midrule 
    \multicolumn{7}{c}{\textbf{Models trained on the filtered set with extractive entity chains only (62.7k/3.5k/3.5k)}} \\
    \midrule 
    \pegasus & 44.10/21.24/35.87 & 65.66/28.69/61.86 & 63.47 & 44.72/21.80/36.94 & 52.87/35.39/46.99 & 40.77 \\
    \frostecp & 44.29/21.26/36.04 & 67.05/29.11/63.02 & 64.69 & 44.02/20.91/36.21 & 49.59/31.16/43.55 & 38.97 \\
    \frostecpp & 44.87/22.05/36.79 & 66.43/30.00/62.48 & \textbf{65.97} & 44.28/21.15/36.59 & 52.17/34.40/46.25 & 39.10 \\
    $\;\;\;$($d \rightarrow c_{\text{drop}};s$) & 44.56/21.80/36.53 & 65.94/29.51/62.04 & 65.08 & 42.93/19.41/35.40 & 47.29/28.27/41.95 & 31.17 \\
    $\;\;\;$($d \rightarrow c_{\text{oracle}};s$)$^{\ast}$ & 49.60/26.64/42.41 & 97.53/53.35/97.51 & 97.41 & 59.80/38.00/54.10 & 97.95/90.57/97.84 & 97.11 \\
    \bottomrule
  \end{tabular}}
  \end{center}
  \vspace{-0.3cm}
  \caption{Performance on XSum summaries when models are trained on the dataset with extractive entity chains only (data filtering, the bottom block) and when novel entities are dropped from the predicted entity chains using the drop-prompt mechanism ($c_{\text{drop}}$). Results with $^{\ast}$ are with oracle entity chain prompts. We report results on the filtered test set with extractive entity chains only (3.5k) and the rest of the test set with novel entities in the targets (7.8k). Best results are bold faced.} 
  \label{table:xsum-control-hall}
  \vspace{-0.3cm}
\end{table*}

\subsection{Controlling Hallucinations}
\label{subsec:faithfulness}

We demonstrate in two ways the planning with entity chains can be useful in mitigating hallucinations in summarization: {\em data filtering} and {\em drop-prompt mechanism} with entity chains.  We focused on the XSum dataset for these experiments.

\paragraph{Data Filtering using Entity Chains.}

During training \frost prepends reference summaries with their entity chains to better plan them. We leverage this to filter the dataset to only keep examples where summaries have fully extractive entity chains; an entity chain is fully extractive if all entities in it can be found in the input document. It's notable that this filtered dataset will not contain novel entities in the summary targets. \frost models trained on this data will ground the summary generation to extracted entity chains while allowing abstraction for non-entity related generations. The resulting XSum dataset has 62.7k/3.5k/3.5k 
train/validation/test instances. We finetune our models from Table~\ref{table:summarization-results} on this filtered dataset and report their performance on the filtered test set (3.5k). We also evaluate them on the rest of the test set (7.8k) where reference entity chains are not fully extractive to their input documents.\footnote{Such data divergence is not unique to XSum, around 30\% of CNN/DailyMail summaries also have reference entity chains that are not fully extractive to their input documents. Writing these summaries requires either document-level inference or the background knowledge of the input documents to generate novel entities or numbers.} See Figure~\ref{fig:frost-predictions} for examples.  

\paragraph{Drop-Prompt Mechanism.}

\frost decoders are trained to generate the summary $s$ following the generation of its plan $c$.
To improve faithfulness, we take the predictions from \frostecpp\footnote{The drop-prompt mechanism can be used with \frostecp also, we have simply used the best among \frostecp and \frostecpp on the XSum set in terms of summary-level \rouge, entity planning and \entfscore.} and modify the generated plan $c$ to $c_{\text{drop}}$ by dropping entities (or parts of them) that are not found in the input document. We then prompt our decoder with  $c_{\text{drop}}$ to generate a new summary. We conduct this with both models, one trained on the full dataset and another on the filtered subset. We also report results for the oracle entity chain prompts $c_{\text{oracle}}$ for a comparison.

\begin{figure}[t!]
  \center{\scriptsize 
  \setlength\tabcolsep{0.1cm}
    \begin{tabular}{p{7.5cm}}
    \toprule 
    \textbf{\gold:} \textcolor{forestgreen}{Walsall} have signed defender \textcolor{orangered}{Luke} \textcolor{forestgreen}{Leahy} on a \textcolor{orangered}{two}-year contract from \textcolor{forestgreen}{Scottish} \textcolor{orangered}{Championship} side \textcolor{forestgreen}{Falkirk}.\\
    \midrule
    \multicolumn{1}{c}{\textbf{Trained on the original XSum dataset}} \\
    \midrule
    \textbf{\pegasus:} \textcolor{forestgreen}{Walsall} have signed defender \textcolor{orangered}{Paddy} \textcolor{forestgreen}{Leahy} from \textcolor{forestgreen}{Scottish} \textcolor{orangered}{Championship} side \textcolor{forestgreen}{Falkirk} on a \textcolor{orangered}{three}-year deal. \\
    
    \textbf{\frostecp:} \textcolor{midnightblue}{[ENTITYCHAIN]} \textcolor{forestgreen}{Walsall} | \textcolor{forestgreen}{Falkirk} | \textcolor{orangered}{Paddy} \textcolor{forestgreen}{Leahy} | \textcolor{orangered}{two} \textcolor{midnightblue}{[SUMMARY]} Walsall have signed \textcolor{forestgreen}{Falkirk} defender \textcolor{orangered}{Paddy} \textcolor{forestgreen}{Leahy} on a \textcolor{orangered}{two}-year deal. \\
    
    \textbf{\frostecpp:} \textcolor{midnightblue}{[ENTITYCHAIN]} \textcolor{forestgreen}{Walsall} | \textcolor{forestgreen}{Falkirk} | \textcolor{orangered}{Liam} \textcolor{forestgreen}{Leahy} | \textcolor{orangered}{two} \textcolor{midnightblue}{[SUMMARY]} \textcolor{forestgreen}{Walsall} have signed \textcolor{forestgreen}{Falkirk} defender \textcolor{orangered}{Liam} \textcolor{forestgreen}{Leahy} on a \textcolor{orangered}{two}-year deal. \\
    
    \textbf{\frostecpp, $c_{\text{drop}}$:} \textcolor{midnightblue}{[ENTITYCHAIN]} \textcolor{forestgreen}{Walsall} | \textcolor{forestgreen}{Falkirk} | \textcolor{forestgreen}{Leahy} \textcolor{midnightblue}{[SUMMARY]} \textcolor{forestgreen}{Walsall} have signed \textcolor{forestgreen}{Falkirk} defender \textcolor{forestgreen}{Leahy} on a free transfer. \\
    
    \midrule
    \multicolumn{1}{c}{\textbf{Trained on the filtered XSum dataset}} \\
    \midrule

    \textbf{\pegasus:}  \textcolor{forestgreen}{Walsall} have signed \textcolor{forestgreen}{Falkirk} defender \textcolor{orangered}{Declan} \textcolor{forestgreen}{Leahy} for an undisclosed fee.
\\
    
    \textbf{\frostecp:} \textcolor{midnightblue}{[ENTITYCHAIN]} \textcolor{forestgreen}{Walsall} | \textcolor{forestgreen}{Falkirk} | \textcolor{orangered}{Paddy} \textcolor{forestgreen}{Leahy} \textcolor{midnightblue}{[SUMMARY]} \textcolor{forestgreen}{Walsall} have signed \textcolor{forestgreen}{Falkirk} defender \textcolor{orangered}{Paddy} \textcolor{forestgreen}{Leahy} for an undisclosed fee. \\
    
    \textbf{\frostecpp:}  \textcolor{midnightblue}{[ENTITYCHAIN]} \textcolor{forestgreen}{Walsall} | \textcolor{forestgreen}{Falkirk} | \textcolor{orangered}{Conor} \textcolor{forestgreen}{Leahy} \textcolor{midnightblue}{[SUMMARY]} \textcolor{forestgreen}{Walsall} have signed \textcolor{forestgreen}{Falkirk} defender \textcolor{orangered}{Conor} \textcolor{forestgreen}{Leahy} for an undisclosed fee. \\
    
    \textbf{\frostecpp, $c_{\text{drop}}$:} \textcolor{midnightblue}{[ENTITYCHAIN]} \textcolor{forestgreen}{Walsall} | \textcolor{forestgreen}{Falkirk} | \textcolor{forestgreen}{Leahy} \textcolor{midnightblue}{[SUMMARY]} \textcolor{forestgreen}{Walsall} have signed \textcolor{forestgreen}{Falkirk} defender \textcolor{forestgreen}{Leahy} for an undisclosed fee. \\
    \bottomrule
    \end{tabular}     
  }
  \vspace{-0.3cm}
  \caption{Example XSum predictions for models presented in Table~\ref{table:xsum-control-hall} and \ref{table:xsum-control-hall-fiathfuleval}. We highlight entities in \textcolor{orangered}{orange} that are not faithful to the input document. Entities in \textcolor{forestgreen}{green} are faithful to the input document. }
  \label{fig:frost-predictions}
  \vspace{-0.5cm}
\end{figure}

\begin{table*}[t!]
  \begin{center}{\tiny 
  \begin{tabular}{ l | c c c c | c c | c c c c | c c } 
    \toprule
    \multirow{3}{*}{\textbf{Model}} & \multicolumn{6}{c|}{\textbf{Test set with Extractive Entity Chains Only (3.5k)}} & \multicolumn{6}{c}{\textbf{Test set with Non-Extractive Entity Chains Only (7.8k)}} \\
    & \multicolumn{4}{c|}{\textbf{Faithfulness}} & \multicolumn{2}{c|}{\textbf{Overall}}  & \multicolumn{4}{c|}{\textbf{Faithfulness}} & \multicolumn{2}{c}{\textbf{Overall}}  \\
    & \textbf{Entail.} & \textbf{\entprec} & \textbf{Human} & \textbf{Agree.} & \textbf{Human} & \textbf{Agree.} & \textbf{Entail.} & \textbf{\entprec} & \textbf{Human} & \textbf{Agree.} & \textbf{Human} & \textbf{Agree.}  \\
    \midrule 
    \multicolumn{13}{c}{\textbf{Models trained on the original XSum dataset (204k/11.3k/11.3k)}} \\
    \midrule 
    \pegasus & 0.613  & 0.800 & 4.20 & 0.74 & 4.09 & 0.69 &  0.402 & 0.361 & 3.15 & 0.71 & 2.93 & \textbf{0.80} \\
    \frostecp &  0.606 & 0.770 & 4.22 & 0.75 & \textbf{4.23} & 0.70  &  0.379 & 0.317 & 3.11 & 0.73 & 2.85 & 0.77 \\
    \frostecpp & 0.589 & 0.751 & 4.13 & 0.72 & 4.11 & 0.66 & 0.371 & 0.357 & 3.31 & \textbf{0.79} & 2.81 & 0.77  \\
    $\;\;\;$($d \rightarrow c_{\text{drop}};s$) & 0.650 & \textbf{0.943} & 4.09 & 0.73 & 4.09 & 0.64 & 0.441 & 0.746 & 3.53 & 0.75 & 3.13 & 0.79 \\
    \midrule 
    \multicolumn{13}{c}{\textbf{Models trained on the filtered set with extractive entity chains only (62.7k/3.5k/3.5k)}} \\
    \midrule 
    \pegasus &  \textbf{0.667} & 0.887 & 4.39 & \textbf{0.80} & 4.14 & 0.69 & 0.389 & 0.501 & 3.37 & \textbf{0.79} & 3.09 &  0.76 \\
    \frostecp & 0.581 & 0.858 & 4.27 & 0.76 & 4.16 & 0.66  & 0.442 & 0.548 & 3.43 & 0.72 & 3.01 & 0.73 \\
    \frostecpp & 0.502 & 0.806 & 4.19 & 0.77 & 4.19 & 0.66  & \textbf{0.465} & 0.491 & 3.55 & 0.76 & 3.16 & 0.76 \\
    $\;\;\;$($d \rightarrow c_{\text{drop}};s$) & 0.533 &  \textbf{0.943} & \textbf{4.41} & 0.75 & 4.18 & \textbf{0.72} & 0.453 & \textbf{0.826} & \textbf{3.85} & 0.74 & \textbf{3.46} & 0.77  \\ 
    \bottomrule
  \end{tabular}}
  \end{center}
  \vspace{-0.3cm}
  \caption{Faithfulness assessment using automatic (Entailment and \entprec) and human evaluations, and overall quality assessment by humans for models presented in Table~\ref{table:xsum-control-hall}. Following \newcite{durmus2020feqa}, agreement (Agree.) is computed by taking the percentage of the annotators that annotate the majority class for the given (document, summary) pair. Best results are bold faced.} 
  \label{table:xsum-control-hall-fiathfuleval}
  \vspace{-0.4cm}
\end{table*}


\paragraph{Effect on Summary-level \rouge, Entity Planning and Specificity.}

Results are presented in Table~\ref{table:xsum-control-hall}. First of all, similar to our results in Table~\ref{table:summarization-results} on the full XSum test set, our results with the models trained on the original dataset further validate that the pretraining and finetuning with the content planning objective is equally useful for both test sets: one where entities are simply copied from the input documents and another where novel entities were inferred, to generate corresponding targets.

The data filtering using entity chains are particularly useful for test cases with extractive reference entity chains; models trained on the filtered data lead to much higher \entfscore (e.g., 65.97 vs 63.03, for \frostecpp) and higher quality entity plans (entitly-level \rouge of 66.43/30.00/62.48 vs 63.73/30.08/60.10 for \frostecpp), without sacrificing the summary quality substantially (summary-level \rouge of 44.87/22.05/36.79 vs 45.08/22.01/36.80 for \frostecpp). Unsurprisingly, these models don't do well on the test set with non-extractive reference entity chains, such examples were not seen during training. 

The drop-prompt mechanism works particularly well for models trained on the full dataset and evaluated on the test set with extractive reference entity chains. For example, the entity planning \rouge scores and \entfscore for \frost models improve from 63.73/30.08/60.10 to 67.10/30.19/63.58, and, 63.03 to 64.41, respectively. Interestingly, we do not observe a significant drop in the summary-level \rouge scores (45.08/22.01/36.80 vs 44.97/21.97/36.77). Basically, our results demonstrate that when models are trained on the noisy data (with data divergence issues, common in summarization datasets \cite{dhingra-etal-2019-handling,maynez-etal-2020-faithfulness}), the drop-prompt mechanism is very effective in generating high quality summaries when entities need to be simply copied from the input documents.

The drop-prompt mechanism doesn't help models when they are already trained on the filtered training set. Also this mechanism is counter intuitive to use for the test set where novel entities need to be inferred to generate targets, we observe drops for models irrespective of how they were trained, either using the full training set or the filtered set.

\paragraph{Effect on Faithfulness.}

The planning objective in \frost itself does not ensure faithfulness; planning is done with the entities in the target, if the target is noisy, its plan may also be noisy (see Figure~\ref{fig:frost-predictions} for predicted entity chains with hallucinated entities). But the planning with the entity chains facilitates the data filtering and the drop-prompt mechanism, using entity chains. Table~\ref{table:xsum-control-hall-fiathfuleval} presents our results assessing them for faithfulness. We randomly selected 50 documents for each of test sets (with extractive or non-extractive reference entity chains) and assessed their summaries from all 8 systems (except with $c_{\text{oracle}}$) from Table~\ref{table:xsum-control-hall}. 


The data filtering using entity chains is extremely useful for improving faithfulness in summaries. We see improvements for all models trained on the filtered set (Table~\ref{table:xsum-control-hall-fiathfuleval}, bottom) compared to their counterparts trained on the full training set (Table~\ref{table:xsum-control-hall-fiathfuleval}, top), when evaluated using \entprec and by humans for faithfulness. We don't observe similar improvements in entailment. Contrary to findings in \cite{maynez-etal-2020-faithfulness}, our results suggest that the entailment scores are not a reliable indicator of faithfulness for documents with extractive reference entity chains. 

The drop-prompt mechanism is also very powerful in improving faithfulness. Irrespective of which training data were used to train the models and which test sets they were evaluated on, we see improvements in entailment scores, \entprec and the human assessment of faithfulness across the board, except for a single case where 
the entailment score slightly drops from 0.465 to  0.453. 
Figure~\ref{fig:frost-predictions} demonstrates how we drop hallucinated entities `\textcolor{orangered}{\em Liam}', `\textcolor{orangered}{\em two}' and `\textcolor{orangered}{\em Conor}' from entity chains to enforce models to generate faithful summaries grounded to the modified entity chains.

We achieve the best performance in terms of \entprec and the human assessment of faithfulness when both the data filtering and the drop-prompt mechanism were used. We achieve 0.943 for \entprec and 4.41 for faithfulness for the test set with extractive entity chains only, and 0.826 for \entprec and 3.85 for faithfulness for the test set with non-extractive entity chains. Humans also found that the predictions from these models were the best in terms of overall summary quality.  

Finally, we carried out pairwise comparisons for human assessments for faithfulness and overall summary quality for all models (using a one-way ANOVA with post-hoc Tukey HSD tests; $p < 0.01$). Interestingly, differences among all model pairs for both faithfulness and overall summary quality are insignificant when evaluated on the test set with extractive reference entity chains only. On the more challenging test set with non-extractive entity chains, \frostecpp trained on the filtered training set and with the drop-prompt mechanism is significantly better than all other models except for (i) \frostecpp trained on the filtered training set but without the drop-prompt mechanism and (ii) \frostecpp trained on the original training set and with the drop-prompt mechanism, for both faithfulness and overall summary quality. All other pairwise differences are insignificant. 



\begin{figure}[t!]
  \center{\fontsize{8}{9}\selectfont 
  \setlength\tabcolsep{0.1cm}
    \begin{tabular}{p{7.5cm}}
    \toprule 
    \textbf{\gold: } \textcolor{midnightblue}{[ENTITYCHAIN]} \textcolor{forestgreen}{BP} | \textcolor{forestgreen}{Lord Browne} | \textcolor{forestgreen}{North Sea}  \textcolor{midnightblue}{[Summary]} Former \textcolor{forestgreen}{BP} chief executive \textcolor{forestgreen}{Lord Browne} has warned \textcolor{forestgreen}{North Sea} oil operators' costs must fall in order for them to compete globally. \\
    \midrule
    \textbf{\frostecpp:} \textcolor{midnightblue}{[ENTITYCHAIN]} \textcolor{forestgreen}{Browne} | \textcolor{forestgreen}{BP}  \textcolor{midnightblue}{[Summary]} Lord \textcolor{forestgreen}{Browne}, the former boss of \textcolor{forestgreen}{BP}, has warned that some oil companies could go out of business if costs continue to rise. \\ 
    \midrule
    \textbf{\frostecpp, $c_{\text{mod}}$, Style diversity}  \\
    \textcolor{midnightblue}{[ENTITYCHAIN]} \textcolor{forestgreen}{BP} | \textcolor{forestgreen}{Browne} \textcolor{midnightblue}{[Summary]} The former boss of \textcolor{forestgreen}{BP}, Lord \textcolor{forestgreen}{Browne}, has warned that some oil companies could go out of business if costs continue to rise. \\
    
    
    \textcolor{midnightblue}{[ENTITYCHAIN]} \textcolor{forestgreen}{BBC} | \textcolor{forestgreen}{BP} | \textcolor{forestgreen}{Browne} \textcolor{midnightblue}{[Summary]} In an interview with the \textcolor{forestgreen}{BBC}, former \textcolor{forestgreen}{BP} chief executive Lord \textcolor{forestgreen}{Browne} has warned the oil and gas industry is in danger of "going to the wall". \\



    
    \midrule
    \textbf{\frostecpp, $c_{\text{mod}}$, Topical diversity}  \\
    
    \textcolor{midnightblue}{[ENTITYCHAIN]} \textcolor{forestgreen}{Brent Crude} \textcolor{midnightblue}{[Summary]} The former boss of the world's biggest oil company has warned that the price of \textcolor{forestgreen}{Brent Crude} will not recover until next year. \\
    
    \textcolor{midnightblue}{[ENTITYCHAIN]} \textcolor{forestgreen}{Brent Crude} | \textcolor{forestgreen}{Unions} \textcolor{midnightblue}{[Summary]} Oil companies will have to cut costs dramatically if they are to survive the fall in the price of \textcolor{forestgreen}{Brent Crude}, according to the former chief executive of the \textcolor{forestgreen}{Unions}. \\

    \bottomrule
    \end{tabular}     
  }
  \vspace{-0.3cm}
  \caption{An example of generating summaries with topical and style diversity using modified entity prompts $c_{\text{mod}}$.}
  \label{fig:frost-predictions-emphasis}
  \vspace{-0.6cm}
\end{figure}

\begin{table*}[t!]
  \begin{center}{\tiny 
  \begin{tabular}{ l | c c c c | c c c c } 
    \toprule
    \multirow{3}{*}{\textbf{Model}} & \multicolumn{4}{c|}{\textbf{XSum}} & \multicolumn{4}{c}{\textbf{CNN/DailyMail}} \\
     & \textbf{Summary} & \textbf{Entity Planning} &  \textbf{Specificity} & \textbf{Avg.} & \textbf{Summary} & \textbf{Entity Planning} &  \textbf{Specificity} & {Avg.} \\ 
     & R1/R2/RL & R1/R2/RL & \entfscore & Length & R1/R2/RL & R1/R2/RL & \entfscore & Length \\ 
    \midrule 
    \pegasus ($d \rightarrow s$) & 47.56/24.87/39.40 & 62.15/39.76/56.36 & 53.48 & 22.27  & 44.05/21.69/40.98 & 61.12/46.46/52.40 & 49.93 & 68.65 \\
    CTRLsum ($k;d \rightarrow s$) & -- & -- & -- & -- & 45.65/22.35/42.50 & 62.19/48.35/52.53 & 51.69 & 74.46 \\
    \frost ($d \rightarrow c;s$) & 47.80/25.06/39.76 & 64.09/41.07/58.18 & 55.49 & 21.09 & 45.11/22.11/42.01 & 64.28/49.86/54.96 & 53.22 & 65.75 \\
    \midrule 
    CTRLsum ($k_{\text{oracle}};d \rightarrow s$)  & -- & -- & -- & -- &  64.72/40.56/61.02 & 78.18/67.39/66.17 & 71.35 & 71.76 \\
    \pegasus ($d;c_{\text{oracle}} \rightarrow s$) & 56.58/34.64/49.87 & 94.48/75.17/93.32 & 93.59 & 22.12 & 56.46/33.62/53.51 & 86.35/80.54/83.76 & 82.41 & 64.84 \\
    \pegasus ($c_{\text{oracle}};d \rightarrow s$) & 57.60/35.62/51.11 & 98.13/79.27/97.58 & 97.90 & 22.13 & 61.66/38.43/58.75 & 97.97/96.11/97.72 & 97.43 & 64.24 \\
    \frost  ($d \rightarrow c_{\text{oracle}};s$) & 58.24/36.47/52.16 & 98.93/80.32/98.91 & 98.59 & 21.80 & 61.85/38.95/59.00 & 98.02/96.19/97.96 & 97.31 & 63.86 \\
    \bottomrule
  \end{tabular}}
  \end{center}
    \vspace{-0.3cm}
  \caption{Comparison of \frost with encoder-guided control summarization. The results in the top block are repeated from Table~\ref{table:summarization-results} for comparison. The bottom block presents oracle results with access to oracle keywords ($k_{\text{oracle}}$) in CTRLSum or oracle entity promts ($c_{\text{oracle}}$) in \frost and \pegasus. $d$ and $s$ stand for the input document and the output summary, respectively. All the results are reported on the full test sets. }  \label{table:comparison-control-sum}
  \vspace{-0.5cm}
\end{table*}

\subsection{Generating Diverse Summaries} 
\label{subsec:diversity}

\frost provides a handle to easily control or manipulate content plan for predicted summaries. In Figure~\ref{fig:frost-predictions}, we saw how we can use this to control entity-level hallucinations, but the strength of \frost models go beyond this. Figure~\ref{fig:frost-predictions-emphasis} shows how \frost models can be effectively used in generating summaries with different entity focus by simply modifying the entity prompt. Future work will focus on how to leverage \frost for synthetic data generation for training improvements.

\subsection{Comparison with Encoder-Guided Control Summarization}
\label{subsec:enc-guided-control}

We compare the decoding strategy in \frost to encoder-guided control summarization systems \cite{he2020ctrlsum,dou2020gsum}. Table~\ref{table:comparison-control-sum} presents our results. In particular, we report on CTRLsum \cite{he2020ctrlsum}; first, a keyword extraction system (BERT-based sequence tagger) is used to extract keywords ($k$) from the input document $d$, and then, the extracted keywords are encoded along with the input document (as $k;d$) to generate the summary. We also finetune \pegasus where the entity chain ($c$) is encoded along with the input document $d$ (as $d;c$ or $c;d$). We only report the oracle results (with $c_{\text{oracle}}$) for these models; like CTRLsum, generating an entity chain $c$ during inference will require training an additional entity chain generator, which is out of scope of this paper.

There are several advantages of \frost-style decoding. Unlike encoder-guided control summarization systems, \frost models can be used in a usual way to generate summaries for input documents without relying on external systems to augment them during inference. Additionally, users can modify the generated entity prompts or provide their desired entity prompts to control the generated summaries (see Section~\ref{subsec:faithfulness} and~\ref{subsec:diversity}). To the best of our knowledge, \frost is the first decoder-prompted control summarization model. 

Our results in Table~\ref{table:comparison-control-sum} show that the prompting the decoder with entity prompts are more effective in generating high-quality and grounded summaries compared to when entity prompts are encoded with the input, especially for abstractive summaries; both \pegasus versions ($d;c_{\text{oracle}}$ and $c_{\text{oracle}};d$) perform inferior to \frost  ($d \rightarrow c_{\text{oracle}};s$) in terms of summary-level \rouge, entity-level \rouge and \entfscore on the XSum dataset. CTRLsum ($k_{\text{oracle}};d \rightarrow s$) achieves better \rouge scores than \frost  ($d \rightarrow c_{\text{oracle}};s$) on CNN/DailyMail (64.72/40.56/61.02 vs 61.85/38.95/59.00). This is not surprising as the oracle keywords $k_{\text{oracle}}$ in CTRLSum retain most words from the summary found in the source document; as such due to the extractive nature of CNN/DailyMail summaries, $k_{\text{oracle}}$ tend to be closer to the surface forms of the reference summaries (\rouge scores between $k_{\text{oracle}}$ and reference summaries are 52.70/17.37/44.37). The difference in \rouge scores narrows down between CTRLsum and \frost to 45.65/22.35/42.50 vs 45.11/22.11/42.01 with automatically extracted keywords. \frost outperforms CTRLsum on entity planning and specificity. It is not clear how well CTRLsum's keyword extraction system will generalize to more abstractive datasets where words in reference summaries are often not found in the source document. 

Finally, our \frost models work as a better microplanner and produce concise summaries than those generated by systems without doing content planning or when its done on the input side. For example, the average lengths of CNN/DailyMail test summaries are 68.65, 74.46, 65.75 and 60.70 tokens for \pegasus, CTRLSum, \frost and humans, respectively. See example predictions comparing CTRLSum and \frost in Figure~\ref{fig:frost-ctrlsum-preds}, Appendix~\ref{app-sec:frost-ctrlsum}.

\section{Conclusion}

In this work, we proposed the use of entity chains, both during pretraining and finetuning, to better plan and ground the generation of abstractive summaries. 
Our approach achieves state-of-the-art \rouge results in XSum and SAMSum, and competitive results in CNN/Dailymail. Compared to other guided summarization models, CTRLSum and GSum, which perform slightly better in CNN/Dailymail, our approach is drastically simpler to implement, trains by augmenting the targets only, utilizes no additional parameters than the baseline \pegasus model, and does not 
rely on any external systems to augment the inputs during inference. We further demonstrate that by modifying the entity chain prompts, we can easily control hallucinations in summaries.


\bibliography{acl2021/anthology,frost}
\bibliographystyle{acl_natbib}



\appendix

\section{Hyperparameters for Pretraining and Finetuning}
\label{app-sec:hyperparameters}

Table~\ref{table:pretrainparameters} and~\ref{table:finetuneparameters} present hyperparameters used for pretraining and finetuning \pegasus and \frost base and large sized models. 

\section{Human Evaluation Instructions}
\label{app-sec:human-eval}

\begin{figure*}[th!]
    \centering
    \begin{tabular}{ c } 
    \includegraphics[scale=0.35]{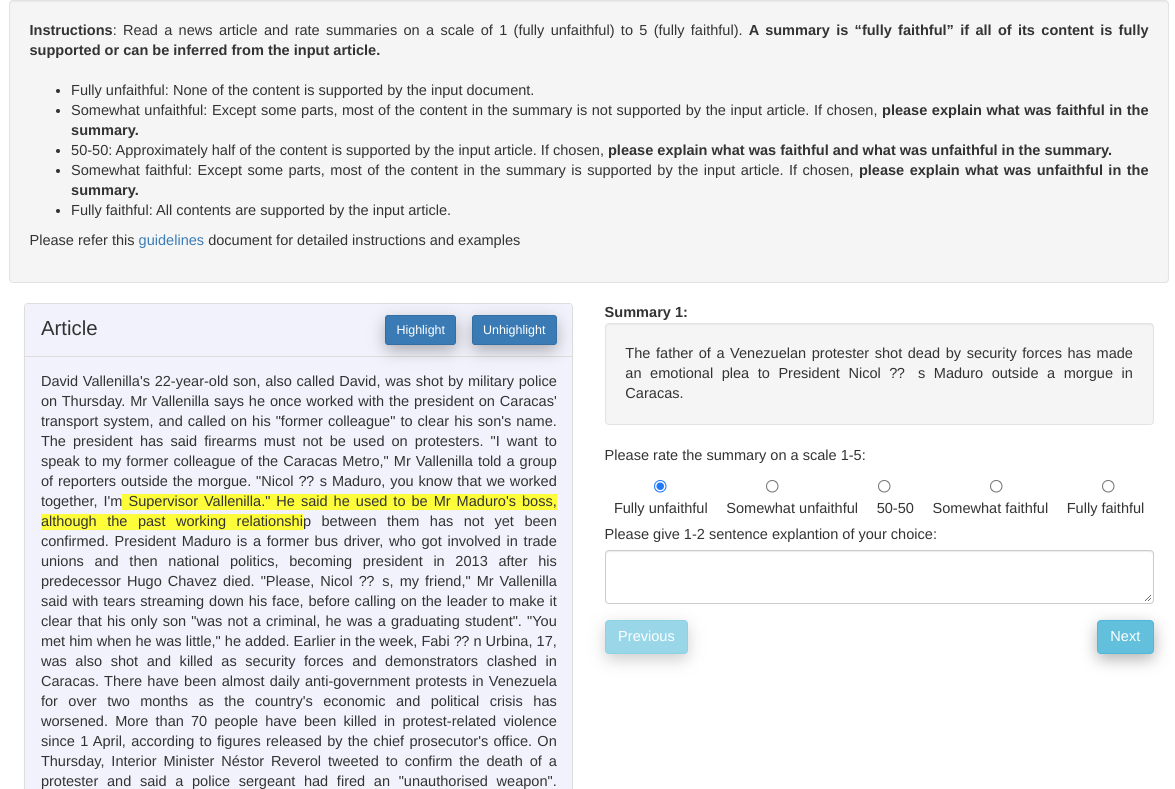}
    \end{tabular}
    \vspace{-0.1cm}
    \caption{Instructions for human evaluations for faithfulness.}
    \label{fig:faithfulness}
    \vspace{-0.2cm}
\end{figure*}

\begin{figure*}[th!]
    \centering
    \begin{tabular}{ c } 
    \includegraphics[scale=0.35]{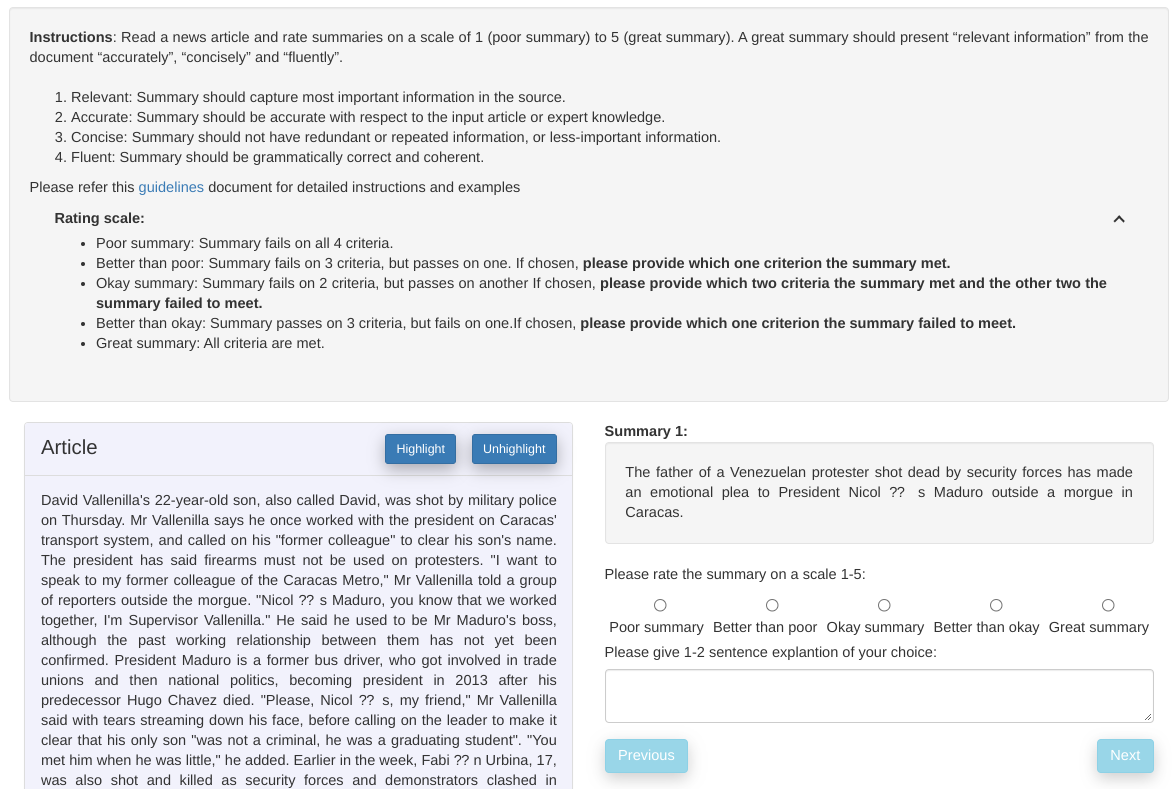}
    \end{tabular}
    \vspace{-0.1cm}
    \caption{Instructions for human evaluations for overall quality of summaries.}
    \label{fig:overall}
    \vspace{-0.2cm}
\end{figure*}

Figure~\ref{fig:faithfulness} and~\ref{fig:overall} show the exact instructions and the template used by our annotators for faithfulness and overall summary quality, respectively. 

\section{Comparison of \frost and CTRLSum}
\label{app-sec:frost-ctrlsum}

Figure~\ref{fig:frost-ctrlsum-preds} presents example predictions from \frost and CTRLSum along with their entity prompts and keywords used for guidance, respectively.

\begin{table*}[th!]
  \begin{center}{\tiny 
  \begin{tabular}{ l | c c c c c c c c} 
    \toprule
    \multirow{2}{*}{Model, Size} & Learning & Label & \multirow{2}{*}{Init.} & \multirow{2}{*}{Steps}  & Batch & \multirow{2}{*}{Corpus} & \multicolumn{2}{c}{Lengths (token)} \\
    & rate & smoothing & & & size & &  input & target \\ \midrule 
    \pegasus, base & 0.01 & 0.1 & random & 1.5m & 1024 & C4/Hugenews & 512 & 256 \\
    \textsc{Frost}(\textsc{F}), base & 0.01 & 0.1 & random & 1.5m & 1024 & C4/Hugenews & 512 & 256\\
    \textsc{Frost}(\textsc{P+F}), base & 0.01 & 0.1 & \pegasus (1m) & 0.5m & 1024 & C4/Hugenews & 512 & 256 \\ \midrule 
    \textsc{Frost}(\textsc{P+F}), large & 0.01 & 0.1 & \pegasus (1.5m) & 1.5m & 1024 & C4/Hugenews & 512 & 256\\
    \bottomrule
  \end{tabular}}
  \end{center}
  \vspace{-0.4cm}
  \caption{Hyperparameters used for pretraining \pegasus and \frost base (223M parameters) and large (568M parameters) models. See text in Section~\ref{subsec:ablations} for more details and how these models were ablated. } \label{table:pretrainparameters}
  \vspace{-0.2cm}
\end{table*}

\begin{table*}[th!]
  \begin{center}{\tiny 
  \begin{tabular}{ l | c c c c c c c c} 
    \toprule
    \multirow{2}{*}{Dataset} & Learning & Label & \multirow{2}{*}{Steps} & Batch & Beam & Beam & \multicolumn{2}{c}{Lengths (token)} \\
    & rate & smoothing & & size & size & alpha & input & target \\ \midrule 
    BillSum & 1e-4 & 0.1 & 200k & 256 & 8 & 0.8 & 1024 & 256  \\
    CNN/DailyMail & 1e-4 & 0.1 & 175k & 256 & 8 & 0.8 & 1024 & 256  \\
    Samsum & 1e-4 & 0.1 & 20k & 256 & 8 & 0.8 & 512 & 128 \\
    XSum & 1e-4 & 0.1 & 100k & 256 & 8 & 0.8 & 512 & 128    \\
    \bottomrule
  \end{tabular}}
  \end{center}
  \vspace{-0.4cm}
  \caption{Hyperparameters used for finetuning \pegasus and \frost models on summarization datasets.} \label{table:finetuneparameters}
  \vspace{-0.2cm}
\end{table*}

\begin{figure*}[t!]
  \center{\tiny 
  \setlength\tabcolsep{0.1cm}
    \begin{tabular}{p{15cm}}
    \toprule 
    \textbf{\gold:} Theia, a bully breed mix, was apparently hit by a car, whacked with a hammer and buried in a field. ``She's a true miracle dog and she deserves a good life,'' says Sara Mellado, who is looking for a home for Theia. \\
    \midrule
    \textbf{CTRLSum: (Oracle keywords)} hit car apparently whacked hammer buried field | dog bully breed mix Theia | true miracle deserves good life \textbf{(Summary)} A bully breed mix named Theia was hit by a car, apparently whacked with a hammer and buried in a field. Four days after her apparent death, the dog managed to stagger to a nearby farm. ``She's a true miracle dog and she deserves a good life,'' the dog's foster mother says. \\
    
    \textbf{\frost: (Oracle Entity Prompt)} Theia ||| Sara Mellado | Theia \textbf{(Summary)} The dog, now named Theia, was apparently hit by a car and buried. Sara Mellado is raising money to help pay for Theia's medical care. \\
    
    \midrule
    \textbf{CTRLSum: (Predicted keywords)} stray pooch Washington State hit car head hammer mercy killing buried field survive | dog Theia | found \textbf{(Summary)} A stray pooch in Washington State was hit by a car and buried in a field. The dog was apparently hit on the head with a hammer in a mercy killing. She managed to survive and was found by a worker who took her to a vet. Theia is now being cared for at a veterinary hospital. \\ 
    \textbf{\frost: (Predicted Entity Prompt)} Washington State ||| Theia ||| \textbf{(Summary)} A stray dog in Washington State was apparently hit by a car and buried. The dog, now named Theia, survived but still needs surgery. A fundraising page has been set up to help pay for her care.\\
    \bottomrule
    \end{tabular}     
  }
  \vspace{-0.1cm}
  \caption{Example predictions from \frost and CTRLSum.}
  \label{fig:frost-ctrlsum-preds}
\end{figure*}


\section{Ablation on Entity Chains with Named Entities, Dates and Numbers}
\label{app-sec:abl-ent-detail}

\begin{figure}[th!]
  \footnotesize
  \center{
\begin{tikzpicture}
    \begin{axis}[
        width  = 0.96*\linewidth,
        height = 5cm,
        major x tick style = transparent,
        ybar=2*\pgflinewidth,
        bar width=7pt,
        ymajorgrids = true,
        ylabel = {Fscores},
        symbolic x coords={RL-Sum,R2-EPlan,\entfscore},
        xtick = data,
        ytick = {40,45,50,55},
        scaled y ticks = false,
        enlarge x limits=0.25,
        ymin=35,
        legend style={at={(0.5,1.01)}, anchor=south,legend columns=-1,font=\tiny}
    ]
        \addplot[style={bblue,fill=bblue,mark=none}]
            coordinates {(RL-Sum, 39.59) (R2-EPlan, 39.88) (\entfscore,53.99)};

        \addplot[style={rred,fill=rred,mark=none}]
             coordinates {(RL-Sum,39.57) (R2-EPlan, 40.89) (\entfscore,55.01)};

        \addplot[style={ggreen,fill=ggreen,mark=none}]
             coordinates {(RL-Sum,39.59) (R2-EPlan, 40.76) (\entfscore,55.09)};

        \addplot[style={ppurple,fill=ppurple,mark=none}]
             coordinates {(RL-Sum,39.44) (R2-EPlan, 41.02) (\entfscore,55.18)};

        \legend{None,NE,NE/D,NE/D/Num}
    \end{axis}
\end{tikzpicture}
  }
  \caption{Ablations on entity chains with named entities, dates and numbers. 
  We report summary-level \rougel (RL-Sum), entity chain-level \rougetwo (R2-EPlan) and \entfscore on the XSum validation set.  Similar observations were made for other measures, see detailed results in Table~\ref{app-table:ablation-ne-d-num}. \label{fig:ablation-ne-d-num}}
\end{figure}
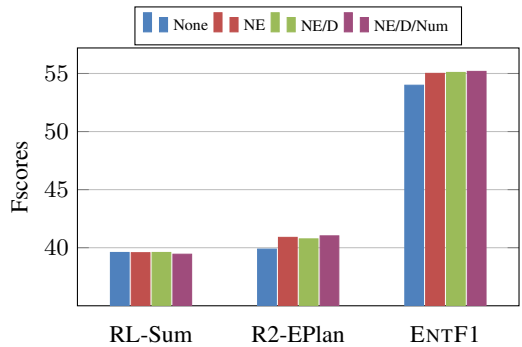

We experimented with entity chains consisting of named entities, dates and numbers. We finetuned \pegasus-large models on the XSum dataset to predict the entity chain and then continue predicting the summary. Results are presented in Figure~\ref{fig:ablation-ne-d-num}. 

We found that the entity chains with all named entities, dates and numbers performed best for entity level measures (entity chain-level \rouge and \entfscore) without sacrificing the final summary quality (summary-level \rouge) significantly. Following this, we always use entity chains with named entities, dates and numbers to have better control over them.

\begin{table*}[t!]
  \begin{center}{\tiny 
  \begin{tabular}{ l | c |  c c c} 
    \toprule
    \multirow{3}{*}{EntityChains} & \multirow{2}{*}{ROUGE} & \multicolumn{3}{c}{Entity Accuracy} \\
    & & Average & Corpus & Total Count (ref=31572) \\
    & R1/R2/RL & R/P/F & R/P/F & found/predicted \\
    \midrule 
    \multicolumn{5}{c}{src $\rightarrow$ tgt} \\
    \midrule 
    None & 47.57/25.05/39.59 & 54.83/60.37/53.99 & 54.17/58.02/56.03 &	17103/29480 \\
    \midrule 
    \multicolumn{5}{c}{src $\rightarrow$ ec$|$tgt} \\
    \midrule 
    NE & 47.58/24.84/39.57 & 55.77/61.06/55.01 & 54.98/59.06/56.94 & 17357/29391 \\
    NE,D & 47.57/24.88/39.59 & 55.56/61.41/55.09 & 54.74/59.59/57.06 &	17284/29007 \\
    NE,D,Num & 47.36/24.69/39.44 & 55.52/61.43/55.18 & 54.73/60.04/57.26 & 17280/28779 \\
    \bottomrule
  \end{tabular}}
  \end{center}
  \caption{Detailed results for ablations on entity chains with named entities, dates and numbers in Figure~\ref{fig:ablation-ne-d-num}. We report results on XSum validation set.} \label{app-table:ablation-ne-d-num}
\end{table*}

\section{Ablation on Sentence-Level Vs Summary-Level Entity Chains}
\label{app-sec:abl-sent-sum}

Table~\ref{app-table:ablation-sum-sent-level} presents detailed results for ablations on sentence-level vs summary-level entity chains in Figure~\ref{fig:ablation-sum-sent-level}. We report results on CNN/DailyMail validation set.

\begin{table*}[t!]
  \begin{center}{\tiny 
  \begin{tabular}{ l | c |  c c c} 
    \toprule
    \multirow{3}{*}{EntityChains} & \multirow{2}{*}{ROUGE} & \multicolumn{3}{c}{Entity Accuracy} \\
    & & Average & Corpus & Total Count (ref=92225) \\
    & R1/R2/RL & R/P/F & R/P/F & found/predicted \\
    \midrule 
    \multicolumn{5}{c}{src $\rightarrow$ tgt} \\
    \midrule 
    None & 45.01/22.57/41.99 &	55.29/51.86/50.66 &	54.17/49.92/51.96 &	49959/100083 \\
    \midrule 
    \multicolumn{5}{c}{src $\rightarrow$ ec$|$tgt} \\
    \midrule 
    Sentence-Level & 44.71/22.51/41.83 & 54.45/54.83/52.06 & 52.87/53.93/53.40 & 48762/90416 \\
    Summary-Level & 45.70/22.61/42.69 &	57.49/54.87/53.82 &	56.08/54.02/55.03 &	51716/95740 \\
    \bottomrule
  \end{tabular}}
  \end{center}
  \caption{Detailed results for ablations on sentence-level vs summary-level entity chains in Figure~\ref{fig:ablation-sum-sent-level}. We report results on CNN/DailyMail validation set.} \label{app-table:ablation-sum-sent-level}
\end{table*}

\begin{table*}[t!]
  \begin{center}{\tiny 
  \begin{tabular}{ l | c c c | c c c } 
    \toprule
    \multirow{3}{*}{\textbf{Model}} & \multicolumn{3}{c|}{\textbf{Test set with Extractive Entity Chains Only (7.8k)}} & \multicolumn{3}{c}{\textbf{Test set with Non-Extractive Entity Chains Only (3.7k)}} \\
    & \textbf{Summary} & \textbf{Entity Planning} & \textbf{Specificity} & \textbf{Summary} & \textbf{Entity Planning} & \textbf{Specificity} \\
    & R1/R2/RL & R1/R2/RL & \entfscore & R1/R2/RL & R1/R2/RL & \entfscore \\
    
    \midrule 
    \multicolumn{7}{c}{\textbf{Models trained on the original CNN/DailyMail dataset (287k/13.4k/11.5k)}} \\
    \midrule 
    \pegasus & 45.09/22.99/42.04 & 61.61/47.88/53.41 & 52.22 & 41.88/18.98/38.77 & 60.10/43.49/50.29 & 45.15 \\
    \frostecp & 45.80/22.96/42.73 & 64.78/51.19/56.02 & 55.49 & 42.87/19.47/39.86 & 63.42/47.05/52.81 & 48.33 \\
    \frostecpp & {46.14}/23.36/43.04 & 64.74/51.21/56.00 & 55.63 & 42.96/19.50/39.86 & 63.32/47.04/52.79 & 48.19 \\
    $\;\;\;$($d \rightarrow c_{\text{drop}};s$) & {46.14}/{23.43}/{43.06} & 64.75/51.18/56.03 & 55.62 & 42.87/19.45/39.76 & 62.96/46.64/52.48 & 47.96 \\
    $\;\;\;$($d \rightarrow c_{\text{oracle}};s$)$^{\ast}$ & 61.56/39.03/58.68 & 98.08/95.71/98.01 & 97.49 & 62.46/38.78/59.67 & 97.89/97.19/97.86 & 96.93 \\
    \midrule 
    \multicolumn{7}{c}{\textbf{Models trained on the filtered set with extractive entity chains only (203.5k/9k/7.8k)}} \\
    \midrule 
    \pegasus & 45.24/22.93/42.13 & 62.12/48.15/53.75 & 52.65 & 41.87/18.79/38.79 & 60.17/43.43/50.23 & 44.96 \\
    \frostecp & 45.76/22.96/42.68 & 64.83/51.02/56.15 & 55.35 & 42.64/19.16/39.50 & 62.67/46.42/52.26 & 47.63 \\
    \frostecpp & 46.11/23.41/43.00 & {65.03}/{51.35}/{56.31} & {55.72} & 42.53/19.12/39.42 & 62.81/46.53/52.42 & 47.69 \\
    $\;\;\;$($d \rightarrow c_{\text{drop}};s$) & 46.07/23.42/42.98 & 65.01/51.32/{56.31} & 55.65 & 42.49/19.13/39.40 & 62.69/46.41/52.33 & 47.65 \\
    $\;\;\;$($d \rightarrow c_{\text{oracle}};s$)$^{\ast}$ & 61.25/38.72/58.32 & 97.88/95.46/97.78 & 97.23 & 61.84/37.92/59.03 & 97.66/96.66/97.53 & 96.15 \\
    \bottomrule
  \end{tabular}}
  \end{center}
  \vspace{-0.3cm}
  \caption{Performance on CNN/DailyMail summaries when models are trained on the dataset with extractive entity chains only (data filtering, the bottom block) and when novel entities are dropped from the predicted entity chains using the drop-prompt mechanism ($c_{\text{drop}}$). Results with $^{\ast}$ are with oracle entity chain prompts. We report results on the filtered test set with extractive entity chains only (7.8k) and the rest of the test set with novel entities in the targets (3.7k).} 
  \label{table:cnndm-control-hall}
  \vspace{-0.2cm}
\end{table*}

\begin{table}[t!]
  \begin{center}{\tiny 
  \begin{tabular}{ l | c c c | c } 
    \toprule
    \multirow{3}{*}{\textbf{Model}} & \multicolumn{4}{c}{\textbf{Test set with Non-Extractive Entity Chains Only (3.7k)}} \\
    & \multicolumn{3}{c|}{\textbf{Faithfulness}} & \textbf{Overall}  \\
    & \textbf{Entail.} & \textbf{\entprec} & \textbf{Human} & \textbf{Human}  \\
    \midrule 
    \multicolumn{5}{c}{\textbf{Trained on the original CNN/DailyMail dataset (204k/11.3k/11.3k)}} \\
    \midrule 
    \pegasus & 0.718 & 0.942 & 4.67 & 4.21 \\
    \frostecp & 0.785 & 0.946 & 4.63 & 4.09 \\
    \frostecpp & 0.721 & 0.907 & 4.63 & 4.12 \\
    $\;\;\;$($d \rightarrow c_{\text{drop}};s$) & {0.798} & 0.966 & 4.63 & 4.21 \\ 
    \midrule 
    \multicolumn{5}{c}{\textbf{Trained on the filtered set with extractive entity chains only (62.7k/3.5k/3.5k)}} \\
    \midrule 
    \pegasus & 0.751 & 0.941 & 4.63 & 4.15 \\
    \frostecp & 0.759 & 0.924 & 4.67 & 4.15 \\
    \frostecpp & 0.732 & 0.903 & 4.61 & 4.10 \\
    $\;\;\;$($d \rightarrow c_{\text{drop}};s$) & {0.798} & 0.972 & 4.71 & 4.18 \\
    \bottomrule
  \end{tabular}}
  \end{center}
  \vspace{-0.3cm}
  \caption{Faithfulness assessment using automatic and human evaluations, and overall quality assessment by humans for models presented in Table~\ref{table:cnndm-control-hall} for the CNN/DailyMail dataset (Test set with Non-Extractive Entity Chains Only).} 
  \label{table:cnndm-control-hall-fiathfuleval}
  \vspace{-0.2cm}
\end{table}

\section{CNN/DailyMail: Controlling Hallucinations Using Entity Chains}
\label{app-sec:hall}

Here, we present results from controlling hallucinations on CNN/DailyMail summaries (i) by training models on the dataset with extractive entity chains only and (ii) by modifying the predicted entity chains to keep only the supported entities. 

Table~\ref{table:cnndm-control-hall} and~\ref{table:cnndm-control-hall-fiathfuleval} present results from these models on the filtered test set with extractive chains only, i.e., 7.8k examples, and on the test set with non-extractive chains only, i.e., 3.7k examples, for CNN/DailyMail. 

Table~\ref{table:xsum-cnndm-control-hall-fulldataset} presents results from these models on the full test sets, i.e., 11.3k examples for XSum and 11.5k examples for CNN/DailyMail.

\begin{table*}[t!]
  \begin{center}{\tiny 
  \begin{tabular}{ l | c c c | c c c } 
    \toprule
    \multirow{3}{*}{Model} & \multicolumn{3}{c|}{\textbf{XSum}} & \multicolumn{3}{c}{\textbf{CNN/DailyMail}} \\
    & Summary & \multicolumn{2}{c|}{Entity Chain} &  Summary & \multicolumn{2}{c}{Entity Chain} \\
    & R1/R2/RL & R1/R2/RL & \entfscore & R1/R2/RL & R1/R2/RL & \entfscore \\
    \midrule 
    & \multicolumn{3}{c|}{\textbf{Trained on Full dataset (204k/11.3k/11.3k)}} & \multicolumn{3}{c}{\textbf{Trained on Full dataset (287k/13.4k/11.5k)}} \\
    \midrule 
    \pegasus (ours) & 47.56/24.87/39.40 & 62.15/39.76/56.36 & 53.48 & 44.05/21.69/40.98 & 61.12/46.46/52.40 & 49.93 \\
    \frostecp & 47.44/24.54/39.24 & 63.57/40.45/57.65 & 54.62 & 44.85/21.83/41.80 & 64.34/49.85/54.98 & 53.17 \\
    \frostecpp & 47.80/25.06/39.76 & 64.09/41.07/58.18 & 55.49 & 45.11/22.11/42.01 & 64.28/49.86/54.96 & 53.22 \\
    $\;\;\;\;\;\;\;\;\;$($d \rightarrow c_{\text{drop}};s$) & 44.94/21.58/37.20 & 54.51/30.46/49.88 & 43.17 & 45.08/22.14/41.99 & 64.17/49.71/54.88 & 53.14 \\
    $\;\;\;\;\;\;\;\;\;$($d \rightarrow c_{\text{oracle}};s$)$^{\ast}$ & 58.24/36.47/52.16 & 98.93/80.32/98.91 & 98.59 & 61.85/38.95/59.00 & 98.02/96.19/97.96 & 97.31 \\
    \midrule 
    & \multicolumn{3}{c|}{\textbf{Trained on the filtered set (62.7k/3.5k/3.5k)}} & \multicolumn{3}{c}{\textbf{Trained on the filtered set (203.5k/9k/7.8k)}} \\
    \midrule 
    \pegasus (ours) & 44.53/21.63/36.61 & 56.82/33.32/51.58 & 47.78 & 44.15/21.59/41.05 &  61.49/46.62/52.61 & 50.16 \\
    \frostecp & 44.10/21.02/36.16 & 54.98/30.53/49.56 & 46.91 & 44.75/21.73/41.65 & 64.13/49.53/54.89 & 52.85 \\
    \frostecpp & 44.46/21.43/36.65 & 56.57/33.04/51.26 & 47.40 & 44.95/22.02/41.84 & 64.31/49.79/55.05 & 53.12 \\
    $\;\;\;\;\;\;\;\;\;$($d \rightarrow c_{\text{drop}};s$) &  43.43/20.15/35.75 & 53.05/28.65/48.15 & 41.64  & 44.91/22.03/41.82 & 64.26/49.73/55.02 & 53.06 \\
    $\;\;\;\;\;\;\;\;\;$($d \rightarrow c_{\text{oracle}};s$)$^{\ast}$ & 56.65/34.49/50.49 & 97.82/79.08/97.74 & 97.20 &  61.44/38.46/58.55 & 97.81/95.85/97.70 & 96.88 \\
    \bottomrule
  \end{tabular}}
  \end{center}
  \caption{Results on the full test sets. Controlling hallucinations by training models on the dataset with extractive entity chains only (the bottom block) and by modifying the predicted entity chains to keep only the supported entities ($c_{\text{drop}}$). Results with $^{\ast}$ are with oracle entity chain prompts. All the results are reported on the full test sets.} 
  \label{table:xsum-cnndm-control-hall-fulldataset}
\end{table*}

\end{document}